\documentclass[10pt,twocolumn,letterpaper]{article}

\usepackage{cvpr}              
\usepackage{algorithm}
\usepackage{algorithmic}
\usepackage{diagbox} 
\usepackage{graphicx}
\usepackage{threeparttable}

\definecolor{cvprblue}{rgb}{0.21,0.49,0.74}
\usepackage[pagebackref,breaklinks,colorlinks,allcolors=cvprblue]
{hyperref}

\def\paperID{6522} 
\def\confName{CVPR}
\def\confYear{2025}

\title{PTDiffusion: Free Lunch for Generating Optical Illusion Hidden Pictures with Phase-Transferred Diffusion Model}

\author{Xiang Gao\quad\quad\quad Shuai Yang\quad\quad\quad Jiaying Liu\thanks{Corresponding author}\\
Wangxuan Institute of Computer Technology, Peking University\\
{\tt\small \{gaoxiang1102, williamyang, liujiaying\}@pku.edu.cn}
}

\begin{document}
\twocolumn[{
\renewcommand\twocolumn[1][]{#1}
\maketitle
\begin{center}
    \captionsetup{type=figure}
    \includegraphics[width=\textwidth]{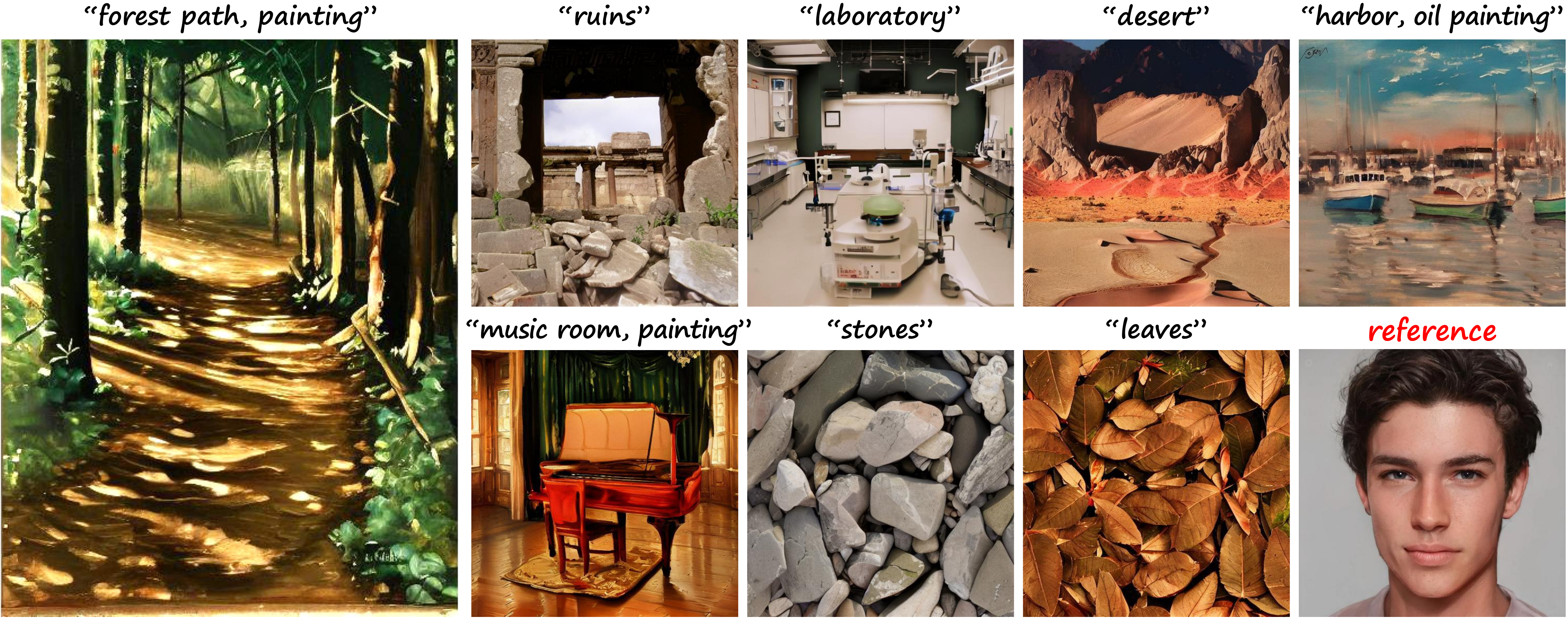}
    \captionof{figure}{Taking the first image on the left as an example, what do you see at your first glance? A painting of a path through a forest (zoom in for a detailed look), or a human face (zoom out for a more global view)? Based on the off-the-shelf text-to-image diffusion model, we contribute a plug-and-play method that naturally dissolves a reference image (shown in the bottom-right corner) into arbitrary scenes described by a text prompt, providing a free lunch for synthesizing optical illusion hidden pictures using diffusion model. Better zoom in.}
\end{center}
}]

\begin{abstract}
Optical illusion hidden picture is an interesting visual perceptual phenomenon where an image is cleverly integrated into another picture in a way that is not immediately obvious to the viewer. Established on the off-the-shelf text-to-image (T2I) diffusion model, we propose a novel training-free text-guided image-to-image (I2I) translation framework dubbed as \textbf{P}hase-\textbf{T}ransferred \textbf{Diffusion} Model (PTDiffusion) for hidden art syntheses. PTDiffusion harmoniously embeds an input reference image into arbitrary scenes described by the text prompts, producing illusion images exhibiting hidden visual cues of the reference image. At the heart of our method is a plug-and-play phase transfer mechanism that dynamically and progressively transplants diffusion features' phase spectrum from the denoising process to reconstruct the reference image into the one to sample the generated illusion image, realizing deep fusion of the reference structural information and the textual semantic information in the diffusion model latent space. Furthermore, we propose asynchronous phase transfer to enable flexible control to the degree of hidden content discernability. Our method bypasses any model training and fine-tuning process, all while substantially outperforming related text-guided I2I methods in image generation quality, text fidelity, visual discernibility, and contextual naturalness for illusion picture synthesis, as demonstrated by extensive qualitative and quantitative experiments. Our project is publically available at \href{https://xianggao1102.github.io/PTDiffusion_webpage/}{this web page}.
\end{abstract}
    
\section{Introduction}
\label{sec:intro}

 As a special form of artistic design, optical illusion hidden picture exploits human visual system's tendency to perceive patterns, shapes, and colors to conceal a secondary image within the intricate details of a primary image. It has wide applications across various fields, such as enhancing aesthetic appeal in fashion design, creating amusing content in digital entertainment, attracting attention in marketing and advertising, improving observation skills in children education, and visual discernment diagnosis in medical treatment. 

 Computationally generating optical illusions is a long-standing challenging task in computer vision and computer graphics. Early methods focus on exploiting how human brains process visual stimuli to generate elementary visual illusions, such as geometric illusion \cite{ehm2011variational}, color illusion \cite{hirsch2020color}, motion illusion \cite{freeman1991motion}, and viewing distance illusion \cite{oliva2006hybrid}. 
 
 More relevant to image processing, Chu~\textit{et al.}~\cite{chu2010camouflage} propose a re-texturing pipeline to synthesize camouflage images, \textit{i.e.}, conceal a foreground image patch into the textures of a background image. Zhang~\textit{et al.}~\cite{zhang2020deep} design a series of optimization functions to synthesize camouflage images from a style transfer perspective \cite{gatys2016image}. Lamdouar~\textit{et al.}~\cite{lamdouar2023making} propose to employ StyleGAN-based generative model~\cite{karras2019style} to synthesize camouflage images in a data-driven manner. 
 
 Since diffusion models \cite{ho2020denoising} revolutionizing the field of generative AI, tremendous attention has been focused on various diffusion-based AIGC applications, among which there are also explorations in illusion picture synthesis. For example, DiffQRCoder \cite{liao2024diffqrcoder} and Text2QR \cite{wu2024text2qr} leverage ControlNet \cite{zhang2023adding} to integrate scannable QR codes into aesthetic pictures. Diffusion Illusions \cite{burgert2024diffusion} employs T2I diffusion model and score distillation sampling \cite{poole2022dreamfusion,wang2023score} to synthesize images with overlay illusions. Visual Anagrams \cite{geng2024visual} merges noises estimated from different views to generate multi-view optical illusions, realizing image appearance change under a certain pixel permutation such as image flip, image rotation, or jigsaw rearrangement.

 In this paper, we pioneer generating optical illusion hidden pictures (we will use ``illusion pictures'' as an abbreviation in the following) from the perspective of text-guided I2I translation, \textit{i.e.}, translating an input reference image into an illusion picture that complies with the text prompt in semantic content while manifesting structural visual cues of the reference image. Our goal differs from the aforementioned optical illusion methods in three aspects: (\romannumeral1) different from camouflage image generation \cite{chu2010camouflage,zhang2020deep,lamdouar2023making} that overemphasizes content concealment, we pursue visual discernibility of both target semantic content and hidden visual cues; (\romannumeral2) unlike synthesizing camouflage image that conceals content into the texture of an existing background image, we expect generating background elements as per the text description; (\romannumeral3) we do not aim at producing transformation-based (flip, rotation, \textit{etc.}) optical illusions like Diffusion Illusions \cite{burgert2024diffusion} and Visual Anagrams \cite{geng2024visual}, but rather focus on seamlessly dissolving a reference image into arbitrary scenes. By contrast, our goal is more methodologically relevant to text-guided I2I \cite{mokady2023null,tumanyan2023plug,parmar2023zero} and controllable T2I \cite{zhang2023adding,mou2024t2i,zhao2023uni} methods. However, since these methods over-bind I2I correlation by explicitly enforcing feature consistency \cite{mokady2023null,tumanyan2023plug,parmar2023zero} or directly training a control network \cite{zhang2023adding,mou2024t2i}, they are less suitable for I2I translation with large semantic deviation (\textit{i.e.,} suffer from structure-semantic conflict issue), and thus tend to generate contextually unnatural results when applied to synthesize illusion pictures. This enlightens us to explore a disentangled image structure representation to relax I2I correlation binding, as well as an appropriate manner to deeply fuse image structure and semantic information along the sampling process.

 Drawing inspiration from digital signal processing that the phase spectrum of an image determines its structural composition, we propose to leverage diffusion features' phase to disentangle image structure and accordingly propose PTDiffusion, a concise and elegant method based on T2I diffusion model that realizes smooth blending of the reference image's structural cues and the text-indicated semantic content in the Latent Diffusion Model (LDM) \cite{rombach2022high} feature space, producing visually appealing illusion pictures in a plug-and-play manner. Specifically, we employ DDIM inversion \cite{song2020denoising} to construct guidance features along a reference image reconstruction trajectory, and progressively transplant the phase of the guidance features into the corresponding features along the text-guided sampling trajectory, such that structural cues of the input reference image are smoothly penetrated into the sampling process of the target image, yielding generation results exhibiting harmonious illusion effects. Besides, we further propose asynchronous phase transfer to flexibly control the structural penetration strength, endowing our method with controllability to the hidden content discernability in the generated illusion picture. Our method is free from model training, fine-tuning, and online optimization, all while demonstrating noticeable strengths in illusion picture synthesis. The contributions of this work are summarized as follows:
 \begin{itemize}
  \item We pioneer generating optical illusion hidden pictures from the perspective of text-guided I2I translation.
  \item We propose a concise and elegant method that realizes deep fusion of image structure and text semantics via dynamic phase manipulation in the LDM feature space, producing contextually harmonious illusion pictures.
  \item We propose asynchronous phase transfer to enable flexible control over the degree of hidden image discernibility.
  \item Our method dispenses with any training and optimization process, providing a free lunch for synthesizing illusion pictures using off-the-shelf T2I diffusion model. 
\end{itemize}

\section{Related work}
\begin{figure*}[t]
    \centering
    \includegraphics[width=0.95\textwidth]{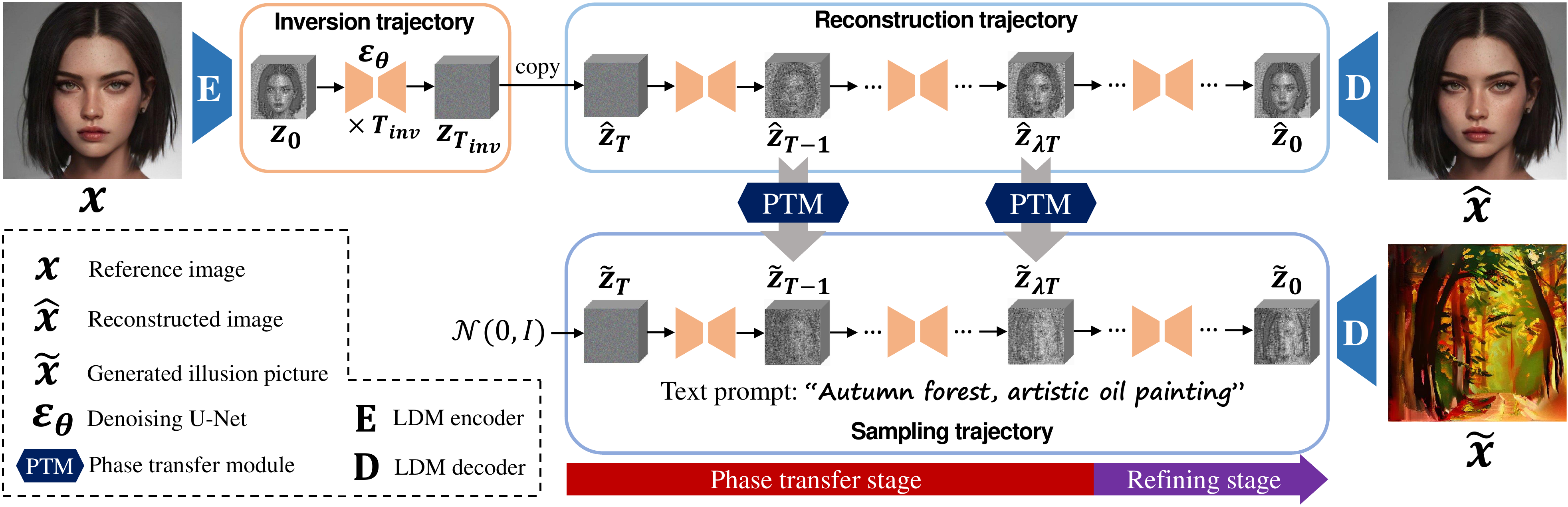}
    \caption{Overview of PTDiffusion. Built upon the pre-trained Latent Diffusion Model (LDM), PTDiffusion is composed of three diffusion trajectories. The inversion trajectory inverts the reference image into the LDM Gaussian noise space. The reconstruction trajectory recovers the reference image from the inverted noise embedding. The sampling trajectory samples the final illusion image from random noise guided by the text prompt. The reconstruction and sampling trajectory are bridged by our proposed phase transfer module, which dynamically transplants diffusion features' phase spectrum to smoothly blend source image structure with textual semantics in the LDM feature space.}
    \label{fig:overall_architecture}
\end{figure*}

\textbf{Text-guided image generation}. Since the advent of DDPM \cite{ho2020denoising}, diffusion model has soon surpassed GAN \cite{goodfellow2020generative} on image synthesis \cite{dhariwal2021diffusion} and has subsequently been accelerated by DDIM \cite{song2020denoising} and extended to conditional image generation paradigm by Palette \cite{saharia2022palette}. After large-scale T2I diffusion models \cite{nichol2022glide,saharia2022photorealistic,ramesh2022hierarchical} remarkably boosting AIGC industry, LDM \cite{rombach2022high} contributes a classical T2I framework with dramatically lowered computational overhead by transferring DDPM from high-dimensional pixel space into low-dimensional feature space, which inspires subsequent T2I models \cite{podell2023sdxl,esser2024scaling} scaling up to larger capacity. To add controllability to T2I synthesis, ControlNet \cite{zhang2023adding} and T2I-Adapter \cite{mou2024t2i} train a control network of the LDM conditioned on certain image priors (edges, depth maps, \textit{etc.}), implicitly realizing reference-image-based structural control to the generated images. However, the over-constraining on object contours and shapes of these methods limit their applicability to synthesizing illusion pictures which emphasize harmonious blending of source image structure and target semantics.

\noindent \textbf{Image-to-image translation}. I2I translation is initially handled by GAN \cite{goodfellow2020generative} in a supervised manner \cite{isola2017image}, and is subsequently extended to unsupervised paradigm through cycle-consistency loss \cite{zhu2017unpaired}, geometry-consistency constraint \cite{fu2019geometry}, contrastive learning \cite{park2020contrastive}, \textit{etc.}, dramatically enriching image stylization applications such as painting stylization \cite{gao2020rpd,kotovenko2019content,gao2025sragan}, font stylization \cite{azadi2018multi,jiang2019scfont}, and image cartoonization \cite{gao2022learning,jiang2023scenimefy}. Since the stunning success of T2I synthesis \cite{nichol2022glide,saharia2022photorealistic,ramesh2022hierarchical,rombach2022high}, attention has been focused on extending large-scale T2I diffusion models to the realm of text-guided I2I translation. SDEdit \cite{meng2021sdedit} translates a reference image by noising it to an intermediate step followed by text-guided denoising. Attention-modulation-based methods such as Null-text inversion \cite{mokady2023null} and pix2pix-zero \cite{parmar2023zero} correlate source and output image by enforcing consistency of cross-attention maps \cite{hertzprompt}. Textual-inversion-based methods like Imagic \cite{kawar2023imagic} and Prompt Tuning Inversion \cite{dong2023prompt} preserve source image visual information via learnable text embedding. Optimization-free methods represented by PAP \cite{tumanyan2023plug} maintain source image structure through dynamic feature injection during the reverse sampling process. Besides, FCDiffusion \cite{gao2024frequency} and FBSDiff \cite{gao2024fbsdiff} leverage DCT frequency bands of diffusion features to control different I2I correlation factors. For illusion picture synthesis, however, these methods struggle to produce contextually natural results with both faithful textual semantics and discernable hidden structure due to the overly bound I2I correlation.

\section{Phase-Transferred Diffusion Model}
The preliminary background of the diffusion model is included in the Supplementary Materials. Below we interpret the overall framework of PTDiffusion, elaborate kernel ingredients, and describe implementation details. 

\subsection{Overall architecture}
As Fig. \ref{fig:overall_architecture} shows, our model builds on the off-the-self LDM \cite{rombach2022high}, and is comprised of an inversion trajectory ($z_{0}\rightarrow z_{T_{inv}}$), a reconstruction trajectory ($z_{T_{inv}}=\hat{z}_{T}\rightarrow \hat{z}_{0}\approx z_{0}$), and a sampling trajectory ($\tilde{z}_{T}\rightarrow \tilde{z}_{0}$). Based on the initial feature $z_{0}=E(x)$ extracted from the reference image $x$ by the LDM encoder $E$, the inversion trajectory adopts a $T_{inv}$-step DDIM inversion \cite{song2020denoising} to project $z_{0}$ into a Gaussian noise $z_{T_{inv}}$ conditioned on the null-text embedding $v_{\emptyset}$:
\begin{equation}
    z_{t+1}=\sqrt{\bar{\alpha}_{t+1}}f_{\theta}(z_{t}, t, v_{\emptyset})+\sqrt{1-\bar{\alpha}_{t+1}}\epsilon_{\theta}(z_{t}, t, v_{\emptyset}),
    \label{eq:dim_inversion}
\end{equation}
\begin{equation}
    f_{\theta}(z_{t}, t, v_{\emptyset})=\frac{z_{t}-\sqrt{1-\bar{\alpha}_{t}}\epsilon_{\theta}(z_{t}, t, v_{\emptyset})}{\sqrt{\bar{\alpha}_{t}}},
    \label{eq:back_to_z0}
\end{equation}
where \{$\bar{\alpha}_{t}$\} are pre-defined DDPM schedule parameters \cite{ho2020denoising}, $\epsilon_{\theta}$ is the LDM denoising U-Net, $f_{\theta}(z_{t}, t, v_{\emptyset})$ is an approximation of $z_{0}$ estimated from $z_{t}$. The reconstruction trajectory applies a $T$-step DDIM sampling to reconstruct $\hat{z}_{0}\approx z_{0}$ from the inverted Gaussian noise $\hat{z}_{T}=z_{T_{inv}}$ conditioned on the same null-text embedding $v_{\emptyset}$:
\begin{equation}
    \hat{z}_{t-1}=\sqrt{\bar{\alpha}_{t-1}}f_{\theta}(\hat{z}_{t}, t, v_{\emptyset}) + \sqrt{1-\bar{\alpha}_{t-1}}\epsilon_{\theta}(\hat{z}_{t}, t, v_{\emptyset}),
    \label{eq:recon}
\end{equation}
where $f_{\theta}(\hat{z}_{t}, t, v_{\emptyset})$ is similar to Eq.~(\ref{eq:back_to_z0}). The sampling trajectory applies a $T$-step DDIM sampling to generate $\tilde{z}_{0}$ from a randomly initialized Gaussian noise $\tilde{z}_{T} \sim \mathcal{N}(0, I)$ conditioned on the text embedding $v$ of the target text prompt. To amplify the influence of text guidance, we exploit classifier-free guidance technique \cite{ho2022classifier} by linearly combining the conditional (target text) and unconditional (null text) noise estimation with a guidance scale $\omega$ at each time step of the sampling trajectory:
\begin{equation}
    \tilde{z}_{t-1}=\sqrt{\bar{\alpha}_{t-1}}f_{\theta}(\tilde{z}_{t}, t, v, v_{\emptyset}) + \sqrt{1-\bar{\alpha}_{t-1}}\epsilon_{\theta}(\tilde{z}_{t}, t, v, v_{\emptyset}),
    \label{eq:sampling}
\end{equation}
\begin{equation}
    f_{\theta}(\tilde{z}_{t}, t, v, v_{\emptyset})=\frac{\tilde{z}_{t}-\sqrt{1-\bar{\alpha}_{t}}\epsilon_{\theta}(\tilde{z}_{t}, t, v, v_{\emptyset})}{\sqrt{\bar{\alpha}_{t}}},
    \label{eq:sampling_z0}
\end{equation}
\begin{equation}
    \epsilon_{\theta}(\tilde{z}_{t}, t, v, v_{\emptyset})=\omega\cdot \epsilon_{\theta}(\tilde{z}_{t}, t, v)+(1-\omega)\cdot\epsilon_{\theta}(\tilde{z}_{t}, t, v_{\emptyset}).
    \label{eq:sampling_classifier_guidance}
\end{equation}

\begin{figure}[t]
    \centering
    \includegraphics[width=3.25in]{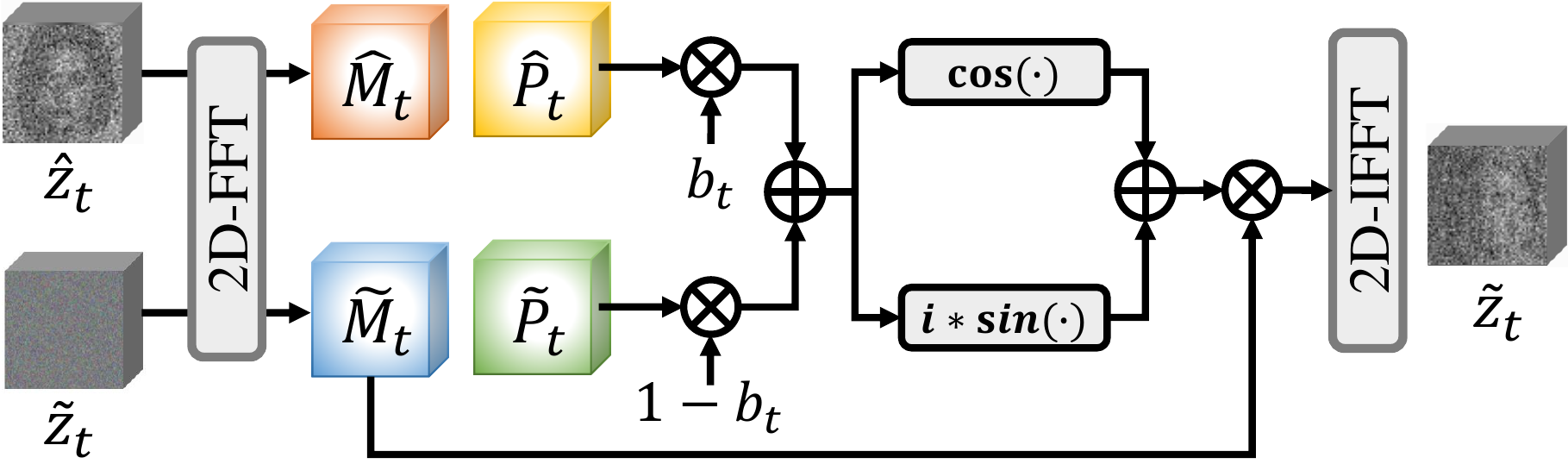}
    \caption{Illustration of the phase transfer module (PTM). To transfer the phase of $\hat{z}_{t}$ into $\tilde{z}_{t}$, we apply 2D FFT to decompose their magnitude $\hat{M}_{t}$, $\tilde{M}_{t}$ and phase $\hat{P}_{t}$, $\tilde{P}_{t}$, linearly fuse their phase with a blending coefficient $b_{t}$, and recombine the fused phase with $\tilde{M}_{t}$. Finally, the manipulated FFT feature is converted back to the spatial domain via 2D IFFT to form the phase-transferred $\tilde{z}_{t}$.}
    \label{fig:PTM}
\end{figure}

To dissolve $x$ into $\tilde{x}$, we propose phase transfer module (PTM) which dynamically blends the structural information of $\hat{z}_{t}$ into $\tilde{z}_{t}$ along the two parallel denoising trajectories. We apply per-step structural penetration realized by PTM only in the early part of the sampling trajectory (which we call phase transfer stage) while leaving the rear part (which we call refining stage) totally unconstrained to guarantee high-quality image synthesis. The two stages are separated by the time step $\lambda T$, where $\lambda$ denotes the proportion of the refining stage to the entire sampling trajectory. The final sampling result $\tilde{z}_{0}$ is transformed to the generated illusion picture via the LDM decoder $D$, \textit{i.e.}, $\tilde{x}=D(\tilde{z}_{0})$.  

\subsection{Phase transfer module}
As the kernel ingredient of our method, the PTM is illustrated in Fig. \ref{fig:PTM}. To transfer the phase of $\hat{z}_{t}$ into the corresponding feature $\tilde{z}_{t}$, we firstly utilize 2D Fast Fourier Transform (FFT) to extract their magnitude and phase:
\begin{equation}
    \hat{R}_{t}+i\hat{I}_{t}=FFT(\hat{z}_{t}), \tilde{R}_{t}+i\tilde{I}_{t}=FFT(\tilde{z}_{t}),
\end{equation}
\begin{equation}
    \hat{M}_{t}=\sqrt{\hat{R}_{t}^{2}+\hat{I}_{t}^{2}}, \tilde{M}_{t}=\sqrt{\tilde{R}_{t}^{2}+\tilde{I}_{t}^{2}},
\end{equation}
\begin{equation}
    \hat{P}_{t}=arctan(\hat{I}_{t}/\hat{R}_{t}),
    \tilde{P}_{t}=arctan(\tilde{I}_{t}/\tilde{R}_{t}),
\end{equation}
where $i$ denotes the imaginary unit, \textit{i.e.}, $i^{2}=-1$. $\hat{R}_{t}$ and $\tilde{R}_{t}$ are the real part of the 2D FFT spectrum of $\hat{z}_{t}$ and $\tilde{z}_{t}$ respectively, $\hat{I}_{t}$ and $\tilde{I}_{t}$ are the corresponding imaginary part. $\hat{M}_{t}$ and $\tilde{M}_{t}$ are the magnitude spectrum of $\hat{z}_{t}$ and $\tilde{z}_{t}$ respectively, $\hat{P}_{t}$ and $\tilde{P}_{t}$ are the corresponding phase spectrum. Then, the extracted phase spectrum $\hat{P}_{t}$ and $\tilde{P}_{t}$ are linearly blended with a time-dependent blending coefficient $b_{t}$, yielding the fused phase spectrum $P_{t}^{fuse}$ as follows:
\begin{equation}
    P_{t}^{fuse}=b_{t} \times \hat{P}_{t}+(1-b_{t}) \times \tilde{P}_{t}.
\end{equation}
The fused phase is recombined with the original magnitude $\tilde{M}_{t}$ before transforming back to the spatial domain with 2D IFFT, finally resulting in the structurally penetrated $\tilde{z}_{t}$:
\begin{equation}
    \tilde{z}_{t}=IFFT(\tilde{M}_{t} \times (\cos(P_{t}^{fuse}) + i\times \sin(P_{t}^{fuse}))).
\end{equation}
Since the structural information of the guidance features \{$\hat{z}_{t}$\} is becoming increasingly prominent as the denoising process proceeds, direct phase transfer in the later denoising steps is prone to harm contextual naturalness of the final result due to excessive structural penetration. Thus, we design a decayed phase blending schedule which gradually decays the blending coefficient \{$b_{t}$\} in the later denoising steps of the phase transfer stage, which we formulate as below:
\begin{equation}
b_{t}=\left\{
\begin{aligned}
1, & & if \ \ \tau T \leq t \leq T \\
1-\sqrt{\frac{\tau T - t}{\tau T - \lambda T}}, & & if \ \ \lambda T \leq t < \tau T
\end{aligned}
\right.
\label{eq:decay_rate}
\end{equation}
where $\lambda \leq \tau \leq 1$. It means that we apply direct phase replacement in the early part of the phase transfer stage where the guidance features $\hat{z}_{t}$ are structurally less distinct, while gradually decaying phase transfer intensity in the later section of the phase transfer stage as $\hat{z}_{t}$ becomes structurally more and more prominent. By decaying phase transfer intensity to avoid overly penetrated structural information, both the visual quality and the contextual naturalness of the generated illusion picture can be noticeably improved.

\subsection{Asynchronous phase transfer}
\begin{figure}[t]
    \centering
    \includegraphics[width=\linewidth]{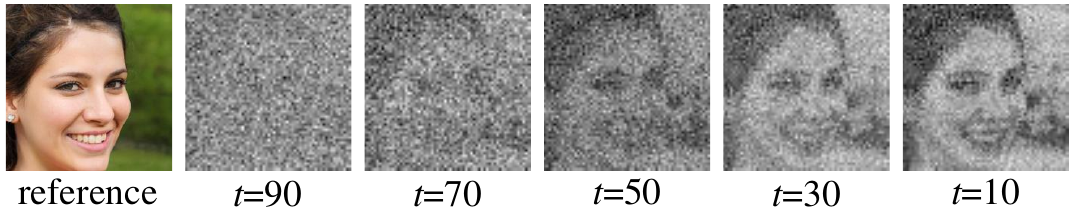}
    \caption{Visualization of the guidance features \{$\hat{z}_{t}$\} along the 100-step reconstruction trajectory. The structural information of $\hat{z}_{t}$ becomes increasingly distinct as the denoising proceeds.}
    \label{fig:asyn_vis}
\end{figure}

\begin{figure}[t]
    \centering
    \includegraphics[width=\linewidth]{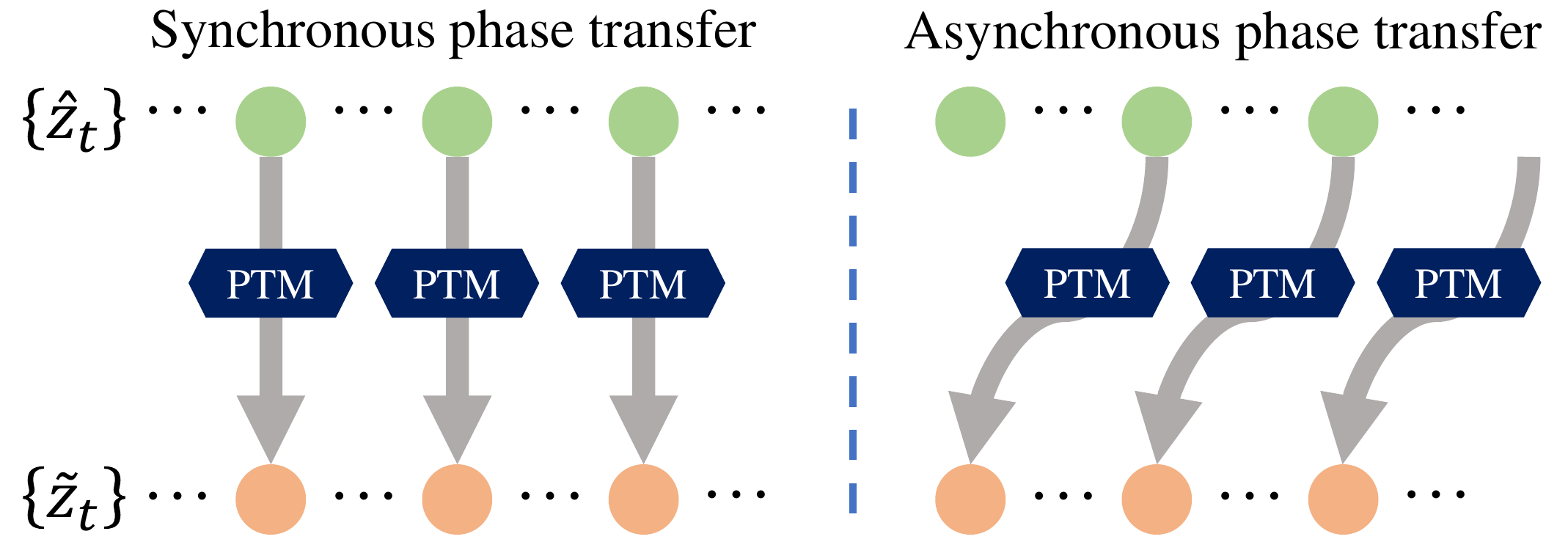}
    \caption{Illustration of the asynchronous phase transfer which transfers phase across diffusion features at different time steps.}
    \label{fig:asyn_illustration}
\end{figure}

Since sufficient denoising steps in the refining stage is of crucial importance to ensure image quality, we fix $\lambda$, namely the length of the phase transfer stage, and propose asynchronous phase transfer to realize controllable structural penetration strength within the fixed-length phase transfer stage. As Fig.~\ref{fig:asyn_vis} displays, later $\hat{z}_{t}$ in the reconstruction trajectory is structurally more prominent than its earlier counterpart. This inspires us to transfer phase from later $\hat{z}_{t}$ with smaller time steps into earlier $\tilde{z}_{t}$ with larger time steps to enhance structural penetration, namely the so-called asynchronous phase transfer as illustrated in Fig.~\ref{fig:asyn_illustration}. 

To this end, we design a concise and elegant solution that implements asynchronous phase transfer based on simple synchronous denoising, \textit{i.e.}, parallel denoising along the reconstruction and sampling trajectory. Specifically, given an async distance $d$, we firstly leverage $\hat{z}_{t}$ to estimate its future counterpart $\hat{z}_{t-d}$ that is $d$ steps ahead of itself in the reconstruction trajectory, then transfer the phase spectrum of the estimated $\hat{z}_{t-d}$ into $\tilde{z}_{t}$. Let APTM denote the asynchronous phase transfer module, we formulate it as follows:
\begin{equation}
\tilde{z}_{t}=APTM(\hat{z}_{t},\tilde{z}_{t},b_{t},d)=PTM(\hat{z}_{t-d}^{*},\tilde{z}_{t},b_{t}),
    \label{eq:aptm}
\end{equation}
where $\hat{z}_{t-d}^{*}$ is a pre-estimation of $\hat{z}_{t-d}$ at time step $t$:
\begin{equation}
    \hat{z}_{t-d}^{*}=\sqrt{\bar{\alpha}_{t-d}}f_{\theta}(\hat{z}_{t}, t, v_{\emptyset}) + \sqrt{1-\bar{\alpha}_{t-d}}\epsilon_{\theta}(\hat{z}_{t}, t, v_{\emptyset}).
\end{equation}
Similar to Eq.~(\ref{eq:back_to_z0}), $f_{\theta}(\hat{z}_{t},t,v_{\emptyset})$ denotes an approximate estimation of $\hat{z}_{0}$ predicted by the current $\hat{z}_{t}$. Note that the async distance $d$ in Eq.~(\ref{eq:aptm}) can also be a negative value to allow for weakened structural penetration strength.
 
\subsection{Implementation details}
We use the pre-trained Stable Diffusion v1.5 as the backbone diffusion model and set the classifier-free guidance scale $\omega$=7.5. To ensure inversion accuracy, we apply 1000-step DDIM inversion for the inversion trajectory, \textit{i.e.}, $T_{inv}$=1000. We apply 100-step DDIM sampling for both the reconstruction and sampling trajectory to save inference time, \textit{i.e.}, $T$=100. During sampling, we allocate 60\% denoising steps to the phase transfer stage and the remaining 40\% steps to the refining stage, \textit{i.e.}, $\lambda$=0.4. We perform direct phase replacement in the early $2/3$ section of the phase transfer stage while performing decayed phase transfer in the later $1/3$ section by setting $\tau$=0.6 in Eq.~(\ref{eq:decay_rate}). The async distance $d$ in Eq.~(\ref{eq:aptm}) is manually tunable (recommended in the range of [-10, 10]) for flexible control over the structural penetration strength, with a default value of $0$. The complete algorithm of PTDiffusion is summarized in Alg. \ref{algorithm}. Experiments are run on a single NVIDIA GeForce RTX 3090 Ti GPU.  

\begin{algorithm}[t]
    \caption{Complete algorithm of PTDiffusion}
    \label{algorithm}
    \begin{algorithmic}[1]
    \renewcommand{\algorithmicrequire}{\textbf{Input:}}
    \renewcommand{\algorithmicensure}{\textbf{Output:}}
    \REQUIRE{Reference image $x$, target text $v$, time steps $T$ and $T_{inv}$, blending coefficients \{$b_{t}$\}, async distance $d$.}
    \ENSURE{The generated illusion picture $\tilde{x}$.}
        \STATE Extract the initial latent feature $z_{0}=E(x)$
        \FOR{$t=0$ to $T_{inv}-1$}
            \STATE Compute $z_{t+1}$ from $z_{t}$ via Eq.~(\ref{eq:dim_inversion})
        \ENDFOR 
        \COMMENT{Inversion trajectory}
        \STATE Initialize $\hat{z}_{T}=z_{T_{inv}}$, $\tilde{z}_{T} \sim \mathcal{N}(0, I)$
        \FOR{$t=T$ to $\lambda T$}
            \STATE Compute $\hat{z}_{t-1}$ from $\hat{z}_{t}$ via Eq.~(\ref{eq:recon})
            \STATE Compute $\tilde{z}_{t-1}$ from $\tilde{z}_{t}$ via Eq.~(\ref{eq:sampling})
            \STATE $\tilde{z}_{t-1}=APTM(\hat{z}_{t-1}, \tilde{z}_{t-1}, b_{t-1}, d)$ as Eq.~(\ref{eq:aptm})
        \ENDFOR\COMMENT{Phase transfer stage}
        \FOR{$t=\lambda T-1$ to $1$}
            \STATE Compute $\tilde{z}_{t-1}$ from $\tilde{z}_{t}$ via Eq.~(\ref{eq:sampling})
        \ENDFOR\COMMENT{Refine stage}
        \STATE Compute the illusion picture $\tilde{x}=D(\tilde{z}_{0})$
    \end{algorithmic}
\end{algorithm}
\section{Experiment}

\begin{figure*}[t]
    \centering
    \includegraphics[width=\linewidth]{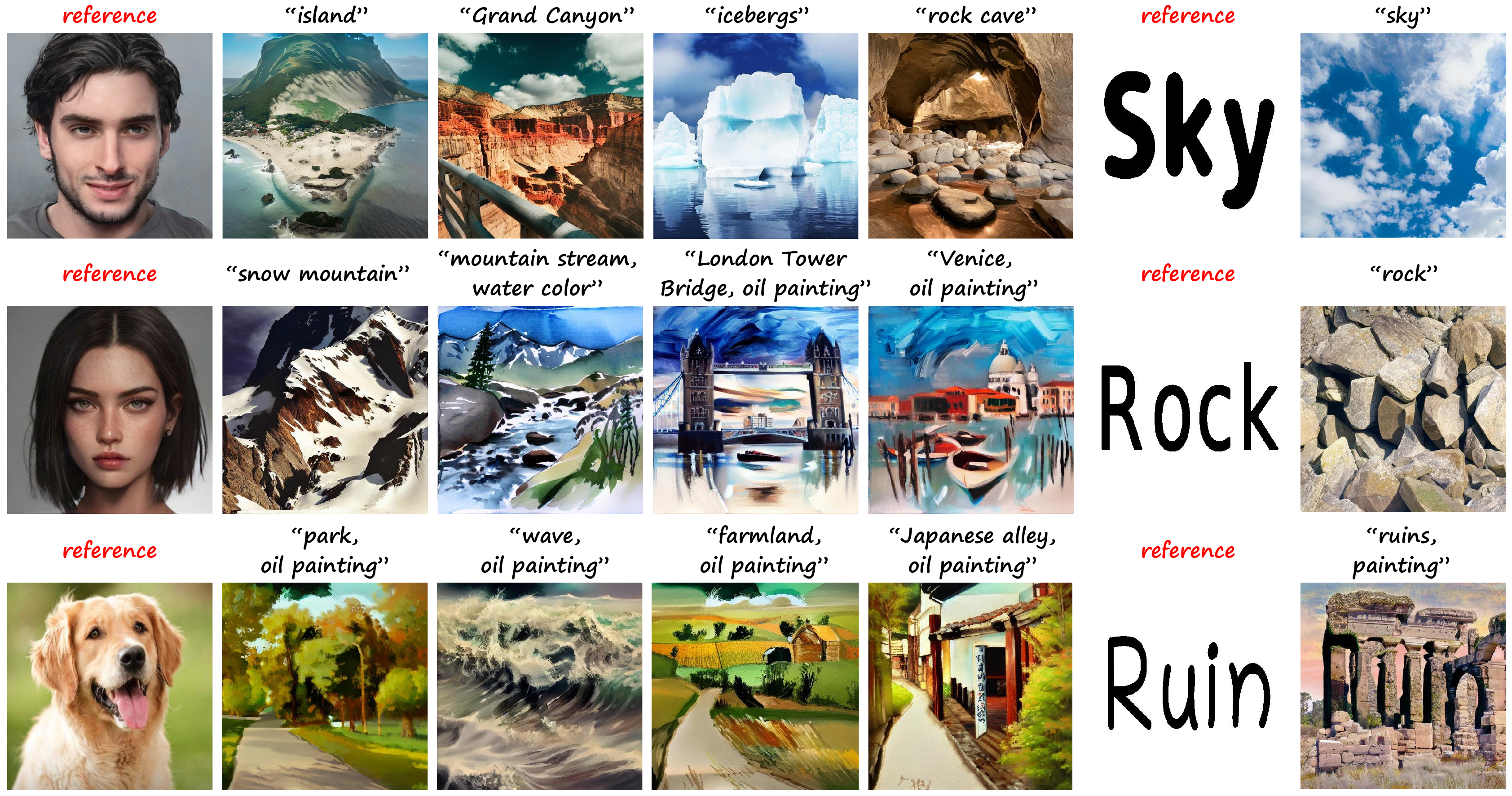}
    \caption{Example results of PTDiffusion in generating optical illusion hidden pictures. Better viewed with zoom-in.}
    \label{fig:qualitative_res}
\end{figure*}

\begin{figure*}[t]
    \centering
    \includegraphics[width=\linewidth]{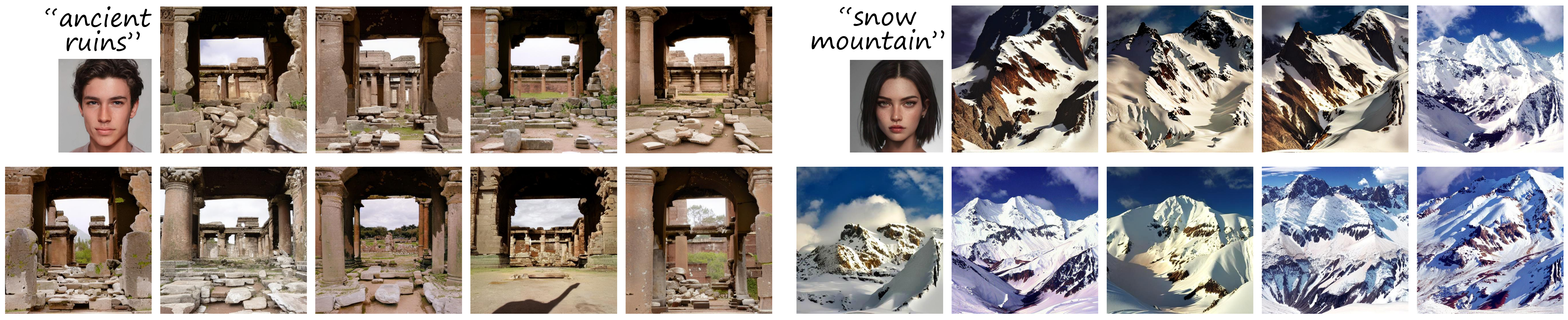}
    \caption{Our method allows to sample diverse illusion pictures with both fixed reference image and text prompt. Better to zoom in.}
    \label{fig:diversity}
\end{figure*}

\begin{figure*}[t]
    \centering
    \includegraphics[width=\textwidth]{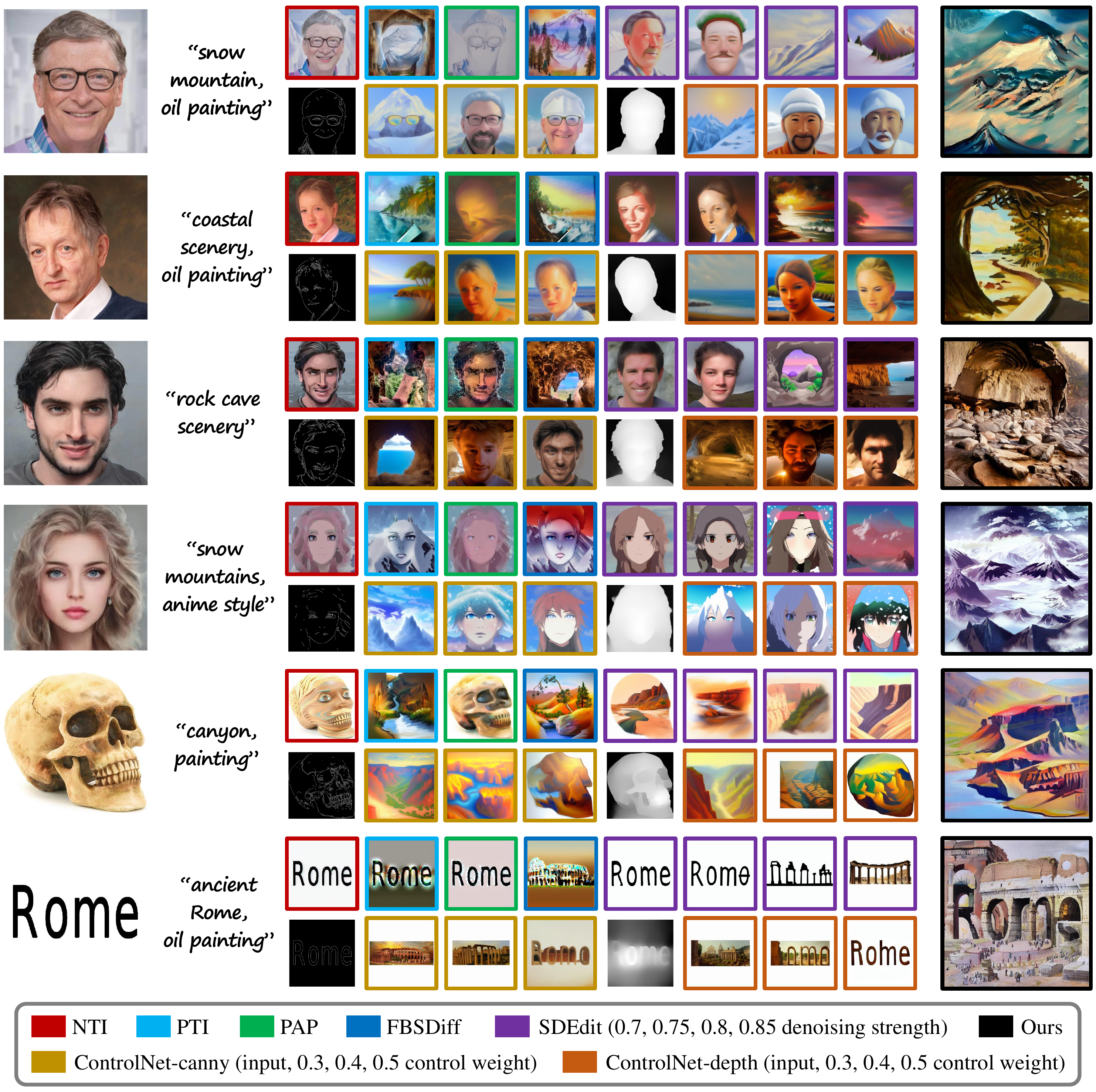}
    \caption{Qualitative comparison to related text-guided I2I and controllable T2I methods on generating optical illusion hidden pictures.}
    \label{fig:method_compare}
\end{figure*}

\begin{figure}[t]
    \centering
    \includegraphics[width=\linewidth]{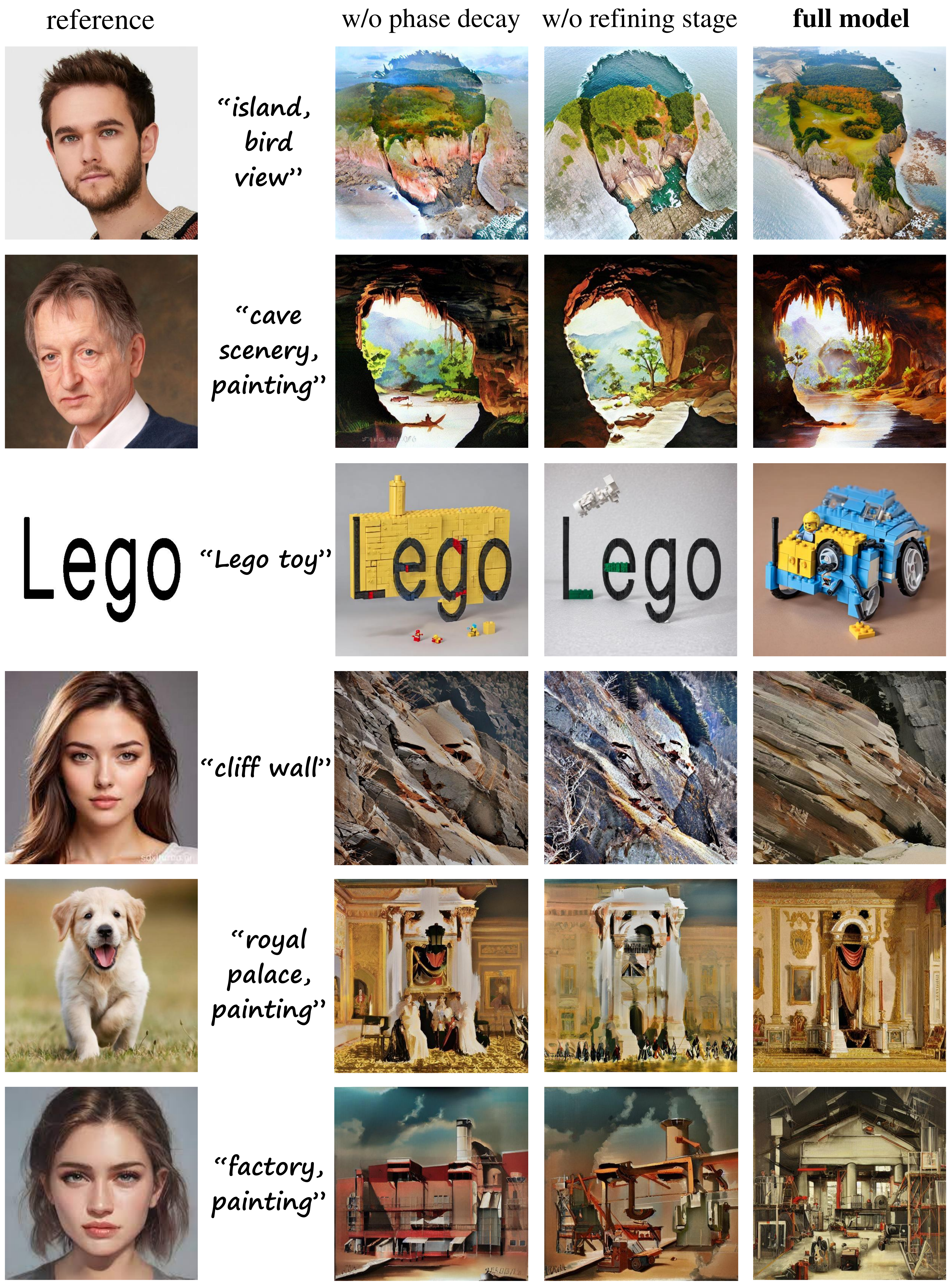}
    \caption{Ablation study about the phase transfer decay and the refining stage of our method. Better viewed with zoom-in.}
    \label{fig:ablation_study}
\end{figure}

\begin{figure}[t]
    \centering
    \includegraphics[width=\linewidth]{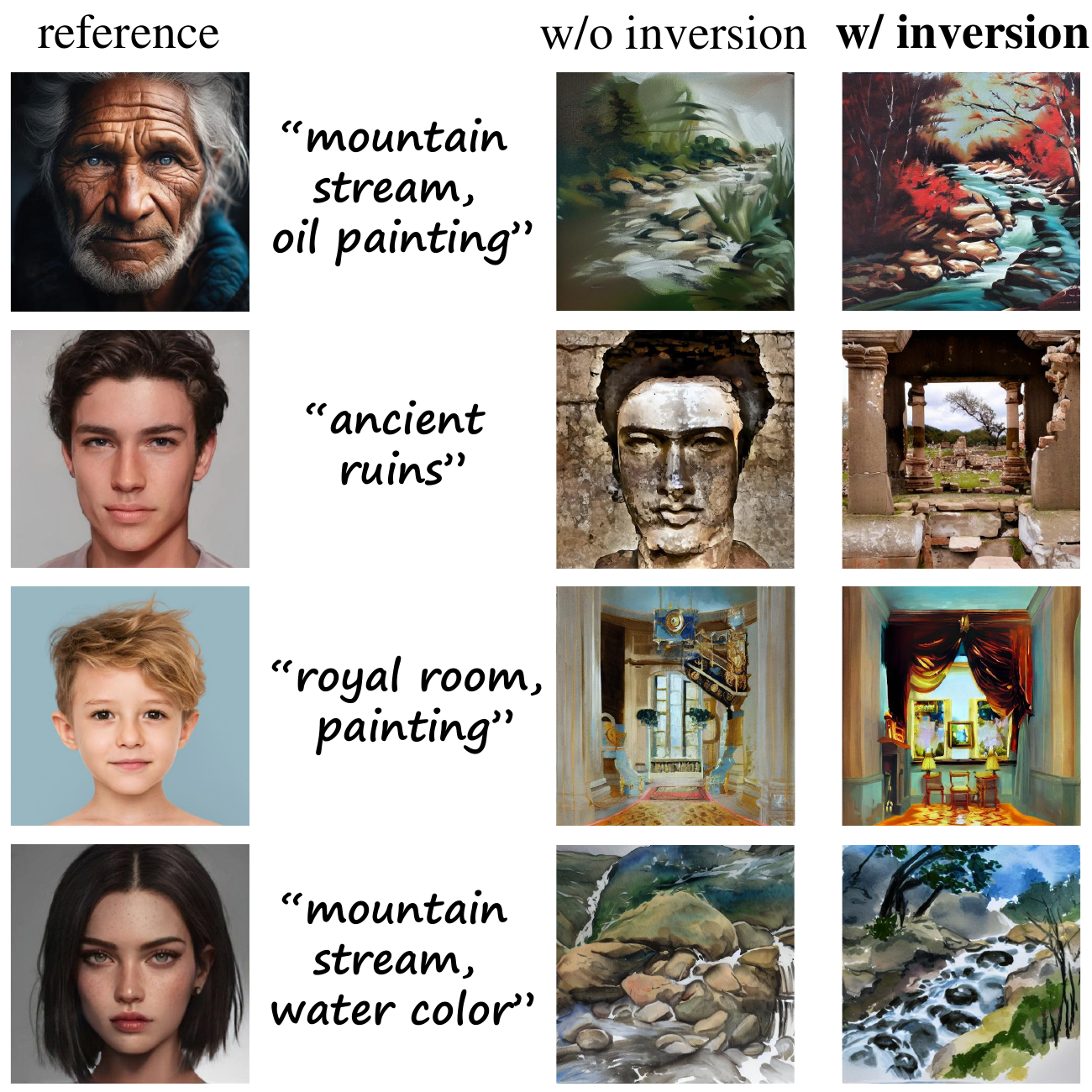}
    \caption{Visual comparison between using DDIM inversion (w/ inversion) and applying forward diffusion process (w/o inversion) to construct the guidance trajectory. Results of using DDIM inversion are more visually appealing and harmonious.}
    \label{fig:inversion_ablation}
\end{figure}

\begin{figure*}[t]
    \centering
    \includegraphics[width=\textwidth]{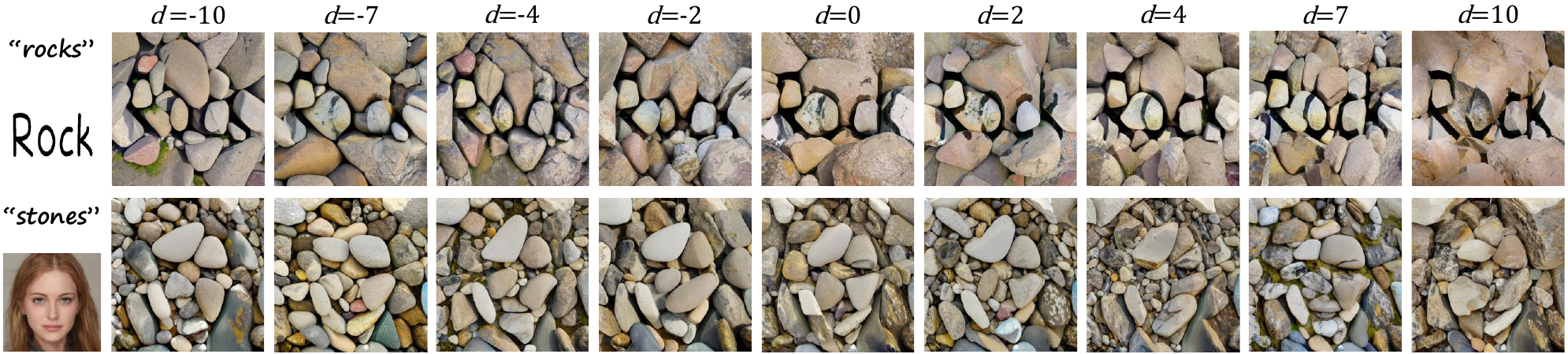}
    \caption{Demonstration of the hidden content discernibility control of our method realized by varying the async distance parameter $d$.}
    \label{fig:APTM}
\end{figure*}

\subsection{Qualitative results}
Example results of our PTDiffusion in producing illusion pictures are displayed in Fig.~\ref{fig:qualitative_res}. Our method harmoniously dissolves a reference image into arbitrary scenes described by a text prompt. Our results are both semantically faithful to the text prompt (zoom in for a detailed look) and structurally perceptible of the hidden reference image (zoom out for a more global view). Apart from real pictures as input reference images, our method also supports synthetic ones as input, for example, integrate a binary text image into a scene of the corresponding semantics to generate contextual text images. It also shows that our method is able to produce high-quality illusion pictures of both realistic domain and artistic domain according to the specific text instructions.

As demonstrated in Fig.~\ref{fig:diversity}, despite using deterministic DDIM inversion and sampling, our method allows to sample diverse results simply by varying the initialized Gaussian noise $\tilde{z}_{T}$, while existing advanced text-guided I2I methods \cite{mokady2023null,parmar2023zero,tumanyan2023plug} do not possess such diversity property.

In Fig.~\ref{fig:method_compare}, we visually compare our method with related text-guided I2I methods including Null-text inversion (NTI) \cite{mokady2023null}, Prompt-tuning inversion (PTI) \cite{dong2023prompt}, PAP \cite{tumanyan2023plug}, FBSDiff \cite{gao2024fbsdiff}, and SDEdit \cite{meng2021sdedit}, as well as controllable T2I method represented by ControlNet \cite{zhang2023adding}. We test different denoising strengths for SDEdit, and test different conditioning modes (Canny edge and depth map) and control weights for ControlNet for comprehensive evaluation of their performance in producing illusion effects. For the remaining methods, we present results with best illusion effect after hyperparameter tuning. NTI produces results with weak target semantics due to the overtight I2I correlation caused by the attention map consistency constraint, which indicates its deficiency in I2I translation with large semantic deviation. Results of PAP are more aligned to the target semantics but still fall short in text fidelity. Besides, it also suffers from the structural overbinding issue due to direct reusage of reference features during the sampling process. Results of PTI and FBSDiff manifest a certain degree of structure-semantic blending, but still underperform in contextual naturalness reflected by visually unpleasant artifacts. Moreover, they suffer from issue of less prominent hidden content, and fail to suit binary text reference images. Results of SDEdit show its difficulty in balancing structure-semantic trade-off, \textit{i.e.}, large denoising strength produces results with overwhelmed structural cues while small one is insufficient to translation input image to the target semantics. Likewise, ControlNet suffers from the same challenge to balance source structure and target semantics when tuning the control weight. By contrast, our PTDiffusion is the only one among the compared methods that realizes harmonious structure-semantic blending, producing visually appealing illusion pictures manifesting both precise textual semantics and clearly discernible hidden content. 

We perform qualitative ablation studies to verify the importance of our method's kernel ingredients. As demonstrated in Fig.~\ref{fig:ablation_study}, both removing phase transfer intensity decay (\textit{i.e.}, apply direct phase replacement along the entire phase transfer stage) and dispensing with the refining stage (\textit{i.e.}, $\lambda = 0$) lead to over-penetrated reference image structure, degrading visual harmony of the generated illusion pictures. In contrast, the full model with partial and decayed phase transfer achieves contextually more natural blending of reference structure and textual semantics.

Our method adopts DDIM inversion to construct the guiding trajectory (the reconstruction trajectory). To verify the necessity of applying DDIM inversion, we have experimented with using the forward diffusion process (\textit{i.e.}, adding noise to $z_{0}$ according to $\hat{z}_{t} \sim \mathcal{N}(\hat{z}_{t}; \sqrt{\bar{\alpha}_{t}}z_{0}, (1-\bar{\alpha}_{t})\mathcal{I})$) to build the guiding trajectory \{$\hat{z}_{t}$\}, which we term the model w/o inversion. Qualitative results (evaluated under the same default hyper-parameters) displayed in Fig. \ref{fig:inversion_ablation} show that the illusion pictures generated by our model w/ inversion are superior to the corresponding results w/o inversion in both visual quality and structure-semantic blending naturalness. This could be due to that the forward diffusion process introduces randomness (Gaussian noise) to the guidance features, yielding unstable guiding trajectory with feature phase irregularly perturbed at each time step, while the guiding trajectory built with DDIM inversion is totally deterministic, providing more stable phase spectra to be transferred along the sampling process.

Fig.~\ref{fig:APTM} qualitatively demonstrates the effectiveness of our proposed asynchronous phase transfer in controlling hidden image discernibility. Increasing async distance $d$ enhances hidden content visual prominence by transferring phase from features at later denoising steps of the reconstruction trajectory into features at earlier denoising steps of the sampling trajectory, while reducing $d$ weakens hidden content discernibility by conversely transferring phase from earlier guidance features into later samping features.

\begin{table}[t]
\renewcommand\arraystretch{1.1}
\fontsize{20}{24}\selectfont  
\begin{center}
\caption{Quantitative evaluations of different text-guided I2I methods for illusion picture synthesis. Results of ControlNet are obtained under the depth condition with 0.4 control weight. Results of SDEdit are evaluated under 0.8 denoising strength.}
\label{tab:quantitative_eval}
    \resizebox{3.25in}{!}{
	\begin{tabular}{c|ccc}
		\hline
		Method & Aesthetic Score ($\uparrow$) & CLIP Score ($\uparrow$) & LPIPS ($\uparrow$)  \\
		\hline
		NTI \cite{mokady2023null} & 6.24 & 0.23 & 0.21  \\
		PTI \cite{dong2023prompt} & 6.18 & 0.28 & 0.52 \\
		PAP \cite{tumanyan2023plug} & 6.09 & 0.25 & 0.40  \\
            FBSDiff \cite{gao2024fbsdiff} & 5.96 & 0.29 & 0.56  \\
            SDEdit \cite{meng2021sdedit} & 6.10 & 0.29 & 0.49  \\
            ControlNet \cite{zhang2023adding} & 6.05 & 0.26 & 0.47  \\
            PTDiffusion (ours) & \bf{6.37} & \bf{0.31} & \bf{0.64} \\
		\hline
	\end{tabular}\vspace{0cm}
    }
\end{center}
\end{table}

\subsection{Quantitative results}
For quantitative evaluations, we measure image visual quality using Aesthetic Score ($\uparrow$) predicted by the pre-trained LAION Aesthetics Predictor V2 model, and measure text fidelity using CLIP Score ($\uparrow$), \textit{i.e.}, the image-text cosine similarity. Considering that weak I2I appearance binding is preferred for generating optical illusion pictures, we measure model's appearance modification ability by evaluating LPIPS ($\uparrow$) between the reference image and the generated illusion image. We test on 80 reference images with each one paired with at least 3 text prompts. Results reported in Tab. \ref{tab:quantitative_eval} demonstrate that our method achieves leading performance for all the aforementioned metrics. 

We conduct quantitative ablation studies about the influence of the phase transfer decay, the refining stage, and the DDIM inversion \textit{w.r.t} the image generation quality and text fidelity. Results displayed in Tab. \ref{tab:quantative ablation} show that missing any one of these ingredients results in declined aesthetic score and image-text CLIP similarity score, which is basically in line with the qualitative results displayed in Fig. \ref{fig:ablation_study} and Fig. \ref{fig:inversion_ablation}. Moreover, the absence of the refining stage causes the most performance drop for these two metrics.  

To quantitatively prove the hidden content perceptibility control ability of our proposed asynchronous phase transfer, we quantify hidden content visual prominence as structure similarity between the input and output image pair, for which we use DINO-ViT self-similarity score \cite{tumanyan2022splicing} as the metric. As reported in Tab. \ref{tab:varying_async}, the I2I structure similarity continuously grows with the increase of the async distance $d$, which tallies with the qualitative results of Fig. \ref{fig:APTM}. It is also worth noting that the increase of $d$ does not lead to drastic drop in CLIP Score, indicating that our proposed asynchronous phase transfer promotes hidden content discernibility without noticeably sacrificing text fidelity.

\begin{table}[t]
\renewcommand\arraystretch{1}
\fontsize{8}{10}\selectfont 
\begin{center}
\caption{Quantitative ablation study \textit{w.r.t} phase intensity decay, refining stage, and DDIM inversion.}
\label{tab:quantative ablation}
    \resizebox{3.25in}{!}{
	\begin{tabular}{c|cc}
		\hline
		Model & Aesthetic Score ($\uparrow$) & CLIP Score ($\uparrow$)\\
		\hline
		  w/o phase decay & 6.24 & 0.28 \\
       
            w/o refining stage & 5.95 & 0.24 \\
            w/o inversion & 6.12 & 0.25 \\
            full model & \textbf{6.37} & \textbf{0.31} \\
		\hline
	\end{tabular}\vspace{0cm}
    }
\end{center}
\end{table}

\begin{table}[t]
\renewcommand\arraystretch{1}
\fontsize{20}{28}\selectfont  
\begin{center}
\caption{Study of the the impact of the async distance $d$ to structural penetration strength and text fidelity.}
\label{tab:varying_async}
    \resizebox{3.25in}{!}{
	\begin{tabular}{c|ccccccc}
		\hline
		Metric & $d$=-9 & $d$=-6 & $d$=-3 & $d$=0 & $d$=3 & $d$=6 & $d$=9\\
		\hline
		  Structure Similarity ($\uparrow$)& 0.880 & 0.893 & 0.902 & 0.908 & 0.912 & 0.917 & 0.926 \\
          
          CLIP Score ($\uparrow$)& 0.309 & 0.311 & 0.305 & 0.307 & 0.301 & 0.298 & 0.293 \\
		\hline
	\end{tabular}\vspace{0cm}
    }
\end{center}
\end{table}

\begin{figure}[t]
    \centering
    \includegraphics[width=\linewidth]{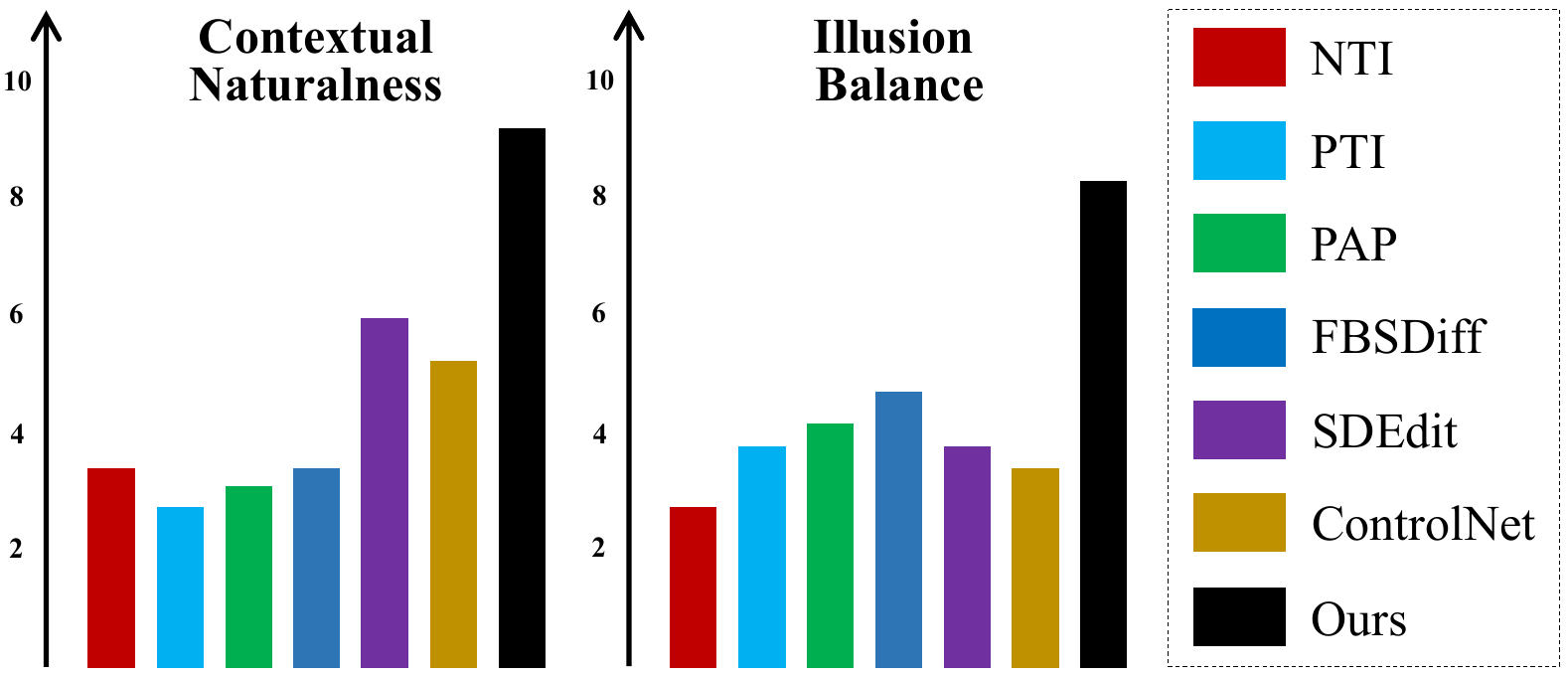}
    \caption{Average user ratings to different methods.}
    \label{fig:user_score}
\end{figure}

For aesthetic assessment of the generated optical illusion hidden pictures, we resort to user study for subjective evaluation. Based on the unique visual characteristics of illusion pictures, we invite 16 participants to score the generation results of different methods on a scale of 1-10 from the following two perspectives: (\romannumeral1) contextual naturalness, \textit{i.e.}, to what extent the reference image is naturally and reasonably blended into the textual scene content; (\romannumeral2) illusion balance, \textit{i.e.}, well-balanced visual discernibility of the textual scene content and the original hidden content, rather than visual prominence of only one side. The average user ratings of all the compared methods in Tab. \ref{tab:quantitative_eval} are reported in Fig. \ref{fig:user_score}, our method outscores other approaches by a large margin in both two dimensions, subjectively indicating significant advantage of our PTDiffusion in illusion picture synthesis.

\section{Conclusion}
This paper pioneers generating optical illusion hidden pictures from the perspective of text-guided I2I translation, \textit{i.e.}, translating a reference image into an illusion picture that is faithful to the text prompt in semantic content while manifesting perceptible structural cues of the reference image. To this end, we propose PTDiffusion, a concise and novel method capable of synthesizing contextually harmonious illusion pictures based on the off-the-shelf T2I diffusion model. At the core of our method is a plug-and-play phase transfer module that smoothly fuses the reference structural information with the target semantic information via progressive phase transfer between the latent diffusion features. We further propose asynchronous phase transfer to enable flexible control over the degree of hidden content discernibility. Our method is optimization-free while showing significant advantages in hidden art synthesis. We will extend our method to more advanced T2I backbone in future work.

\clearpage
\setcounter{page}{1}
\maketitlesupplementary

\section{Preliminary background}
\subsection{Diffusion model background}
Since the advent of Denoising Diffusion Probabilistic Model (DDPM), diffusion model has soon dominated research field of generative AI due to its advantages in training stability and sampling diversity as compared with GAN. Grounded in the theory of stochastic differential equations, diffusion model learns to iteratively denoise a noise-corrupted input signal (\textit{e.g.}, an image or a video clip), ultimately generating clean data that follow the underlying target distribution. Diffusion model is conceptually composed of a forward diffusion process and a reverse denoising process. The forward diffusion process gradually adds noise to the data over a series of steps, transforming the data into a random Gaussian distribution, while the reverse denoising process learns to reverse the forward process by iteratively removing noise from the data, starting from pure noise and gradually reconstructing the original data. The model is trained to predict the noise added at each step of the forward process. By learning to denoise, the model can generate new data samples by starting from random noise and applying the reverse process.

Given the original data distribution $q(x_{0})$, the forward diffusion process applies a $T$-step Markov chain to gradually add noise to the original data $x_{0}$ according to the conditional distribution $q(x_{t}|x_{t-1})$, which is defined as follows:
\begin{equation}
    q(x_{t}|x_{t-1})=\mathcal{N}(x_{t}; \sqrt{\alpha_{t}}x_{t-1}, (1-\alpha_{t})\mathcal{I}),
\end{equation}
where $\alpha_{t}$ follows a predefined schedule, $\alpha_{t}\in(0, 1)$, $\alpha_{t} > \alpha_{t+1}$. Using the notation $\bar{\alpha}_{t}=\prod_{i=1}^{t}\alpha_{i}$, we can derive the marginal distribution $q(x_{t}|x_{0})$ that can be used to directly obtain $x_{t}$ from $x_{0}$ in a single step for arbitrary time step $t$:
\begin{equation}
    q(x_{t}|x_{0})=\mathcal{N}(x_{t}; \sqrt{\bar{\alpha}_{t}}x_{0}, (1-\bar{\alpha}_{t})\mathcal{I}),
    \label{xt|x0}
\end{equation}
where $\sqrt{\bar{\alpha}_{T}} \approx 0$. With the forward diffusion process, the source data $x_{0}$ is transformed into $x_{T}$ that follows an isotropic Gaussian distribution.

The reverse denoising process learns to conversely convert a Gaussian noise $x_{T}$ to the manifold of the original data distribution $q(x_{0})$ by gradually estimating and sampling from the posterior distribution $p(x_{t-1}|x_{t})$. Since the posterior distribution $p(x_{t-1}|x_{t})$ is mathematically intractable, we can derive the conditional posterior distribution $p(x_{t-1}|x_{t}, x_{0})$ with the Bayes formula and some algebraic manipulation:
\begin{equation}
    p(x_{t-1}|x_{t},x_{0})=\mathcal{N}(x_{t-1}; \tilde{\mu}_{t}(x_{t}, x_{0}), \tilde{\beta}_{t}\mathcal{I}),
    \label{conditional_posterior}
\end{equation}
\begin{equation}
    \tilde{\mu}_{t}(x_{t}, x_{0})=\frac{\sqrt{\bar{\alpha}_{t-1}}\beta_{t}}{1-\bar{\alpha}_{t}}x_{0}+\frac{\sqrt{\alpha}_{t}(1-\bar{\alpha}_{t-1})}{1-\bar{\alpha}_{t}}x_{t},
\end{equation}
\begin{equation}
    \tilde{\beta}_{t}=\frac{1-\bar{\alpha}_{t-1}}{1-\bar{\alpha}_{t}}\beta_{t},
\end{equation}
where $\beta_{t}=1-\alpha_{t}$. 
However, the conditional posterior distribution $p(x_{t-1}|x_{t}, x_{0})$ cannot be directly used for sampling since $x_{0}$ is unavailable at inference time ($x_{0}$ is the target of the sampling process). Thus, DDPM tries to estimate the unknown $x_{0}$ given the $x_{t}$ at each time step. Considering the reparameterization form of Eq. \ref{xt|x0}:
\begin{equation}
    x_{t}=\sqrt{\bar{\alpha}_{t}}x_{0}+\sqrt{1-\bar{\alpha}_{t}}\epsilon_{t},
    \label{reparam}
\end{equation}
in which $\epsilon_{t}$ denotes the randomly sampled Gaussian noise that maps $x_{0}$ to $x_{t}$ in a single step according to Eq. \ref{xt|x0}. Given Eq. \ref{reparam}, we can represent $x_{0}$ using $x_{t}$ and $\epsilon_{t}$:
\begin{equation}
    x_{0}=\frac{1}{\sqrt{\bar{\alpha}_{t}}}(x_{t}-\sqrt{1-\bar{\alpha}_{t}}\epsilon_{t}).
    \label{repre_x0_with_xt}
\end{equation}
However, the Gaussian noise $\epsilon_{t}$ sampled in the forward diffusion process is also unknown for the reverse denoising process where only $x_{t}$ is available. Consequently, DDPM builds a noise estimation network $\epsilon_{\theta}$ that predicts the sampled Gaussian noise $\epsilon_{t}$ in Eq. \ref{repre_x0_with_xt} with $x_{t}$ and time step $t$ as input, which is realized by training $\epsilon_{\theta}$ with the following noise regression loss:
\begin{equation}
    L=\|\epsilon_{t}-\epsilon_{\theta}(x_{t}, t)\|_{2},
    \label{DDPM_loss}
\end{equation}
where $t\sim$ Uniform($\{1,...,T\}$), $\epsilon_{t} \sim \mathcal{N}(0, \mathcal{I})$, $x_{t}$ is computed via Eq. \ref{reparam}. After model training, $y_{\theta}(x_{t})$, the estimation of $x_{0}$ given $x_{t}$, can be obtained simply by replacing $\epsilon_{t}$ in Eq. \ref{repre_x0_with_xt} with the predicted noise $\epsilon_{\theta}(x_{t}, t)$:
\begin{equation}
    y_{\theta}(x_{t})=\frac{1}{\sqrt{\bar{\alpha}_{t}}}(x_{t}-\sqrt{1-\bar{\alpha}_{t}}\epsilon_{\theta}(x_{t}, t)).
    \label{x_0_pred}
\end{equation}
Replacing the unknown $x_{0}$ in Eq. \ref{conditional_posterior} with its predicted estimation $y_{\theta}(x_{t})$ given by Eq. \ref{x_0_pred}, we can sample $x_{t-1}$ based on $x_{t}$ from the approximate posterior distribution $\mathcal{N}(x_{t-1};\tilde{\mu}_{t}(x_{t},y_{\theta}(x_{t})),\tilde{\beta}_{t}\mathcal{I})$, and thus sample the ultimate $x_{0}$ step by step from the initial Gaussian noise $x_{T}$. 

\subsection{Conditional diffusion model}
Taking the image generation task as an example, conditional diffusion model tackles conditional image synthesis by introducing additional condition $c$ to the model to guide image generation (denoising) process. In this paradigm, the condition signal $c$ together with $x_{t}$ and time step $t$ are taken as input to the noise estimation network $\epsilon_{\theta}$, such that $\epsilon_{\theta}$ is trained to conditionally predict the added Gaussian noise in the forward diffusion process, as supervised by the randomly sampled $\epsilon_{t}$ in Eq. \ref{reparam}. The training loss given by Eq. \ref{DDPM_loss} is correspondingly updated as:
\begin{equation}
    L=\|\epsilon_{t}-\epsilon_{\theta}(x_{t}, t, c)\|_{2},
    \label{conditional_DDPM_loss}
\end{equation}
where $t\sim$ Uniform($\{1,...,T\}$), $\epsilon_{t} \sim \mathcal{N}(0, \mathcal{I})$, $x_{t}$ is computed via Eq. \ref{reparam}. After model training, the reverse sampling process is applied to generate new images from random Gaussian noise $x_{T}$, based on the step-by-step denoising according to the conditional posterior distribution given by Eq. \ref{conditional_posterior}, in which the unknown $x_{0}$ is approximated by the linear combination of $x_{t}$ and the conditional noise estimation, \textit{i.e.}, the $y_{\theta}(x_{t})$ (the approximate $x_{0}$ estimated by $x_{t}$) in Eq. \ref{x_0_pred} is updated as:
\begin{equation}
    y_{\theta}(x_{t}, c)=\frac{1}{\sqrt{\bar{\alpha}_{t}}}(x_{t}-\sqrt{1-\bar{\alpha}_{t}}\epsilon_{\theta}(x_{t}, t, c)).
\end{equation}

\subsection{Denoising diffusion implicit model}
Denoising diffusion implicit model (DDIM) is a variant of diffusion model that builds on the framework of DDPM but enables much more efficient sampling while maintaining high-quality generation results. DDIM can generate samples in significantly fewer steps compared with DDPM by modeling the reverse denoising process as a non-Markovian process and skipping the intermediate denoising steps. 

DDIM is totally the same as DDPM in model training and only differs with DDPM in model inference, namely that DDIM can directly inherit the pre-trained DDPM model. To compute $x_{t-1}$ from $x_{t}$ in the reverse denoising (sampling) process, DDIM features a two-step deterministic denoising. In the first step, DDIM estimates an approximate $x_{0}$ based on $x_{t}$ using Eq. \ref{x_0_pred}. In the second step, DDIM computes $x_{t-1}$ from the approximate $x_{0}$ using the forward diffusion in the form of Eq. \ref{reparam}:
\begin{equation}
    x_{t-1}=\sqrt{\bar{\alpha}_{t-1}}y_{\theta}(x_{t})+\sqrt{1-\bar{\alpha}_{t-1}}\epsilon_{t-1}, 
    \label{ddim_1}
\end{equation}
where $y_{\theta}(x_{t})$ is given by Eq. \ref{x_0_pred}. Considering that the $\epsilon_{t-1}$ in the above equation is the sampled Gaussian noise in the forward diffusion process, which is unknown in the reverse denoising process, we can replace $\epsilon_{t-1}$ with $\epsilon_{\theta}(x_{t-1}, t-1)$, the approximate $\epsilon_{t-1}$ estimated by the network $\epsilon_{\theta}$. Therefore, the Eq. \ref{ddim_1} can be updated as:
\begin{equation}
    x_{t-1}=\sqrt{\bar{\alpha}_{t-1}}y_{\theta}(x_{t})+\sqrt{1-\bar{\alpha}_{t-1}}\epsilon_{\theta}(x_{t-1}, t-1). 
    \label{ddim_2}
\end{equation}
However, the $\epsilon_{\theta}(x_{t-1}, t-1)$ in the above equation is also unavailable since $x_{t-1}$ is unknown (we only know $x_{t}$ and want to compute $x_{t-1}$). Thus, we can further approximate $\epsilon_{\theta}(x_{t-1}, t-1)$ with $\epsilon_{\theta}(x_{t}, t)$ and arrive to the final DDIM sampling equation:
\begin{equation}
    x_{t-1}=\sqrt{\bar{\alpha}_{t-1}}y_{\theta}(x_{t})+\sqrt{1-\bar{\alpha}_{t-1}}\epsilon_{\theta}(x_{t}, t). 
    \label{ddim_3}
\end{equation}
Eq. \ref{ddim_3} shows that the reverse sampling process of DDIM is totally deterministic, namely, each starting Gaussian noise $x_{T}$ yields a unique sampling result $x_{0}$. 

Note that the above derived two-step sampling process of $x_{t} \rightarrow x_{0} \rightarrow x_{t-1}$ also applies for $x_{t} \rightarrow x_{0} \rightarrow x_{t+1}$. That is, a clean image $x_{0}$ can be deterministically inverted into a Gaussian noise through the following inversion process:
\begin{equation}
    x_{t+1}=\sqrt{\bar{\alpha}_{t+1}}y_{\theta}(x_{t})+\sqrt{1-\bar{\alpha}_{t+1}}\epsilon_{\theta}(x_{t}, t). 
    \label{ddim_inversion}
\end{equation}
The DDIM inversion given by Eq. \ref{ddim_inversion} has wide applications in image editing and style transfer. For conditional image generation of DDIM, the $y_{\theta}(x_{t})$ and $\epsilon_{\theta}(x_{t}, t)$ in Eq. \ref{ddim_3} and Eq. \ref{ddim_inversion} are updated to $y_{\theta}(x_{t}, c)$ and $\epsilon_{\theta}(x_{t}, t, c)$ respectively.

\subsection{Latent diffusion model}
Latent diffusion model (LDM) compresses images from high-dimensional pixel space into low-dimensional feature space via pre-trained autoencoder, and builds diffusion model based on the latent feature space, such that computational overhead for both training and inference can be dramatically lowered. The training of LDM is similar to Eq. \ref{conditional_DDPM_loss} except that we use notation $z$ to denote latent features:
\begin{equation}
    L=\|\epsilon_{t}-\epsilon_{\theta}(z_{t}, t, c)\|_{2},
    \label{LDM_loss}
\end{equation}
where $\epsilon_{t} \sim \mathcal{N}(0, \mathcal{I})$, $z_{t}=\sqrt{\bar{\alpha}_{t}}z_{0}+\sqrt{1-\bar{\alpha}_{t}}\epsilon_{t}$, $z_{0}=E(x_{0})$, $E$ is the pre-trained image encoder. The reverse denoising process from $z_{T} \sim \mathcal{N}(0, \mathcal{I})$ to $z_{0}$ is the same as $x_{T} \sim \mathcal{N}(0, \mathcal{I})$ to $x_{0}$ in DDPM. After reverse denoising process, the denoised clean features $z_{0}$ is decoded by the pre-trained decoder $D$ to yield the finally generated image $x_{0}$, \textit{i.e.}, $x_{0}=D(z_{0})$. In LDM framework, the condition $c$ could be the extracted image features that are concatenated with $x_{t}$ as the input of $\epsilon_{\theta}$ for image-to-image translation applications, and also could be the encoded textual features that are interacted with $x_{t}$ with cross-attention layers inside $\epsilon_{\theta}$ for text-to-image synthesis task.

\section{More qualitative results}
\begin{figure*}[t]
    \centering
    \includegraphics[width=0.95\linewidth]{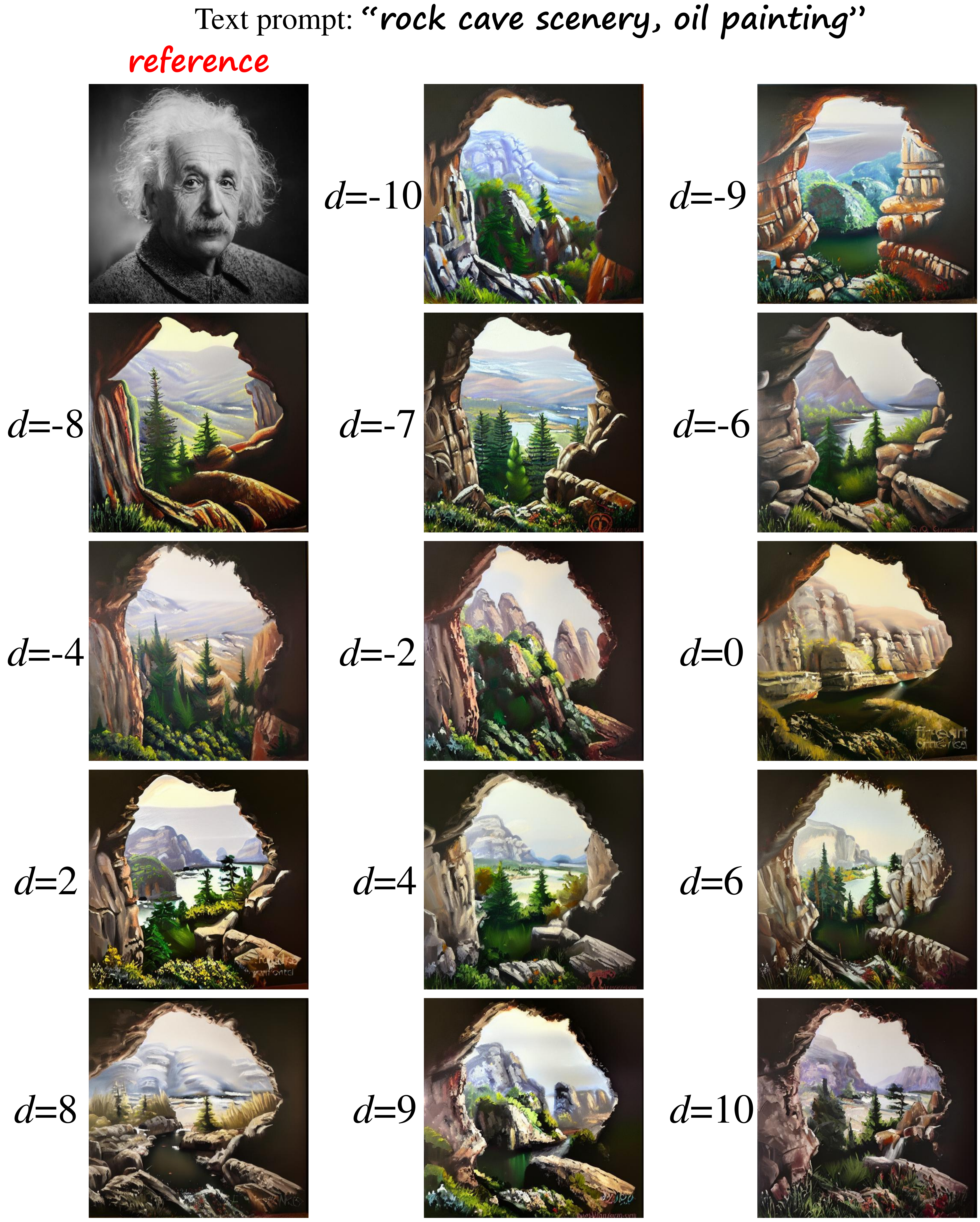}
    \caption{More results of hidden content discernibility control realized by varying the async distance parameter $d$ in our method.}
    \label{fig:async_distance_1}
\end{figure*}

\begin{figure*}[t]
    \centering
    \includegraphics[width=0.95\linewidth]{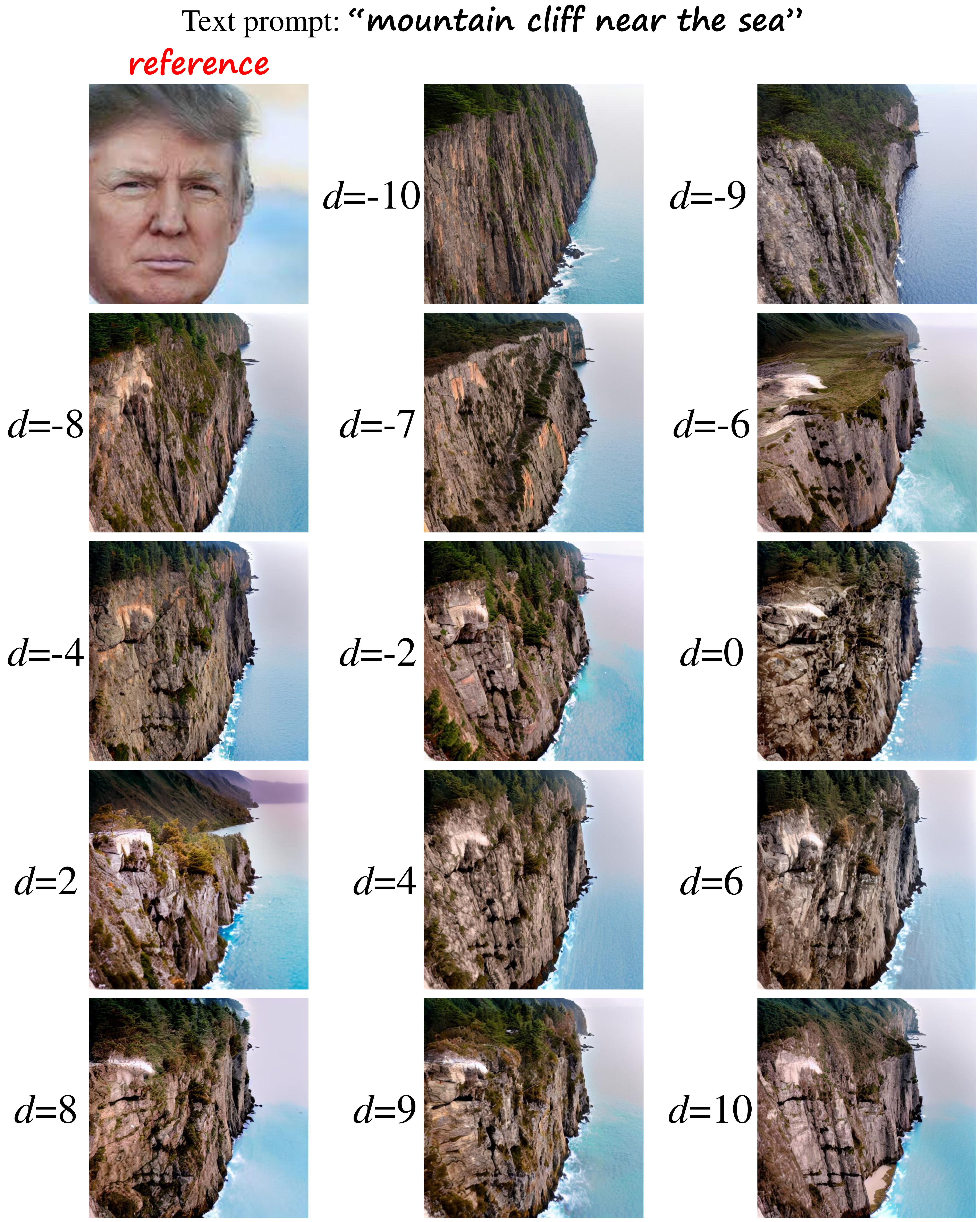}
    \caption{More results of hidden content discernibility control realized by varying the async distance parameter $d$ in our method.}
    \label{fig:async_distance_2}
\end{figure*}

\begin{figure*}[t]
    \centering
    \includegraphics[width=0.95\linewidth]{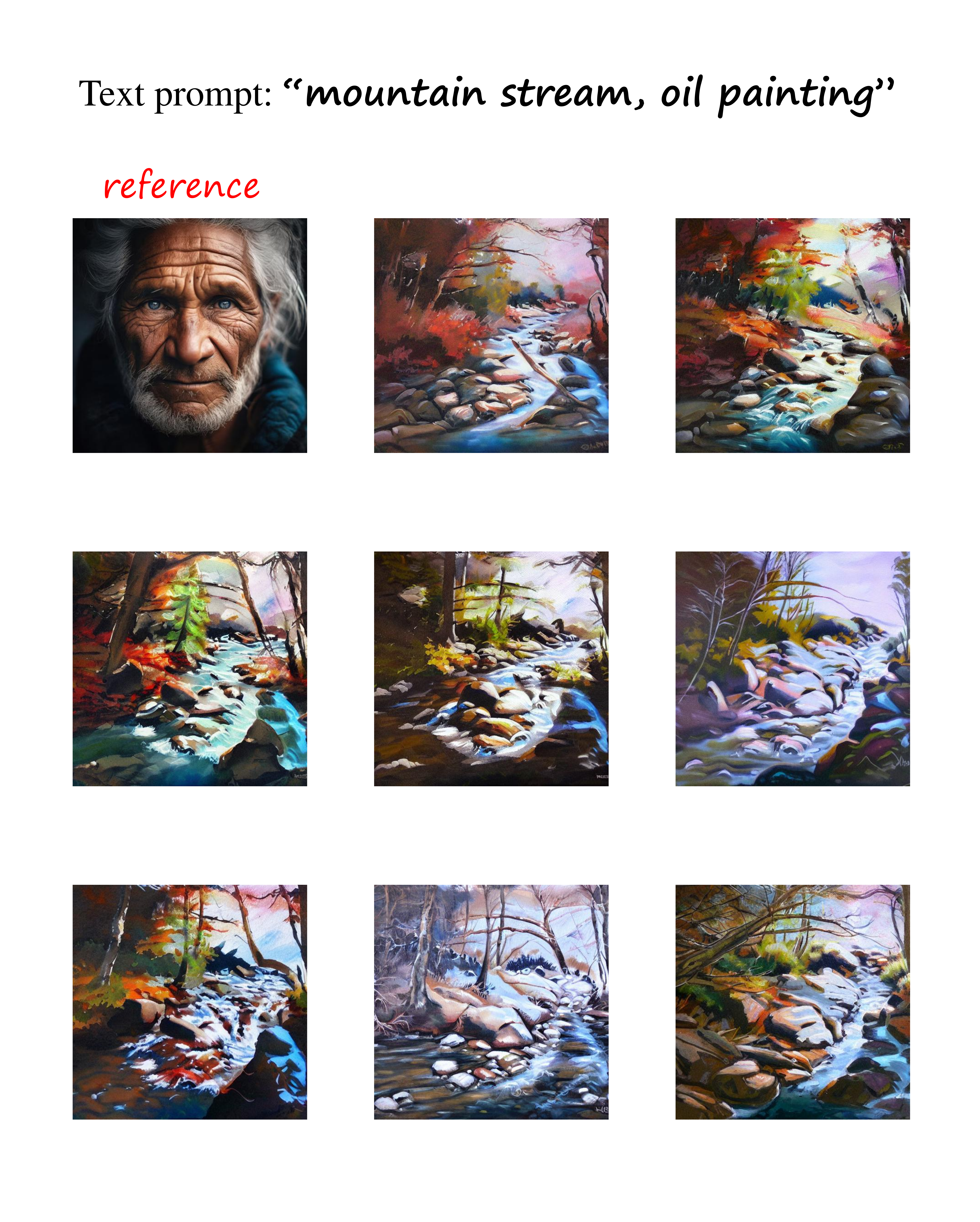}
    \caption{More examples of diversified sampling results of our method realized by varying the initial Gaussian noise $\tilde{z}_{T}$.}
    \label{fig:diverse_1}
\end{figure*}

\begin{figure*}[t]
    \centering
    \includegraphics[width=0.95\linewidth]{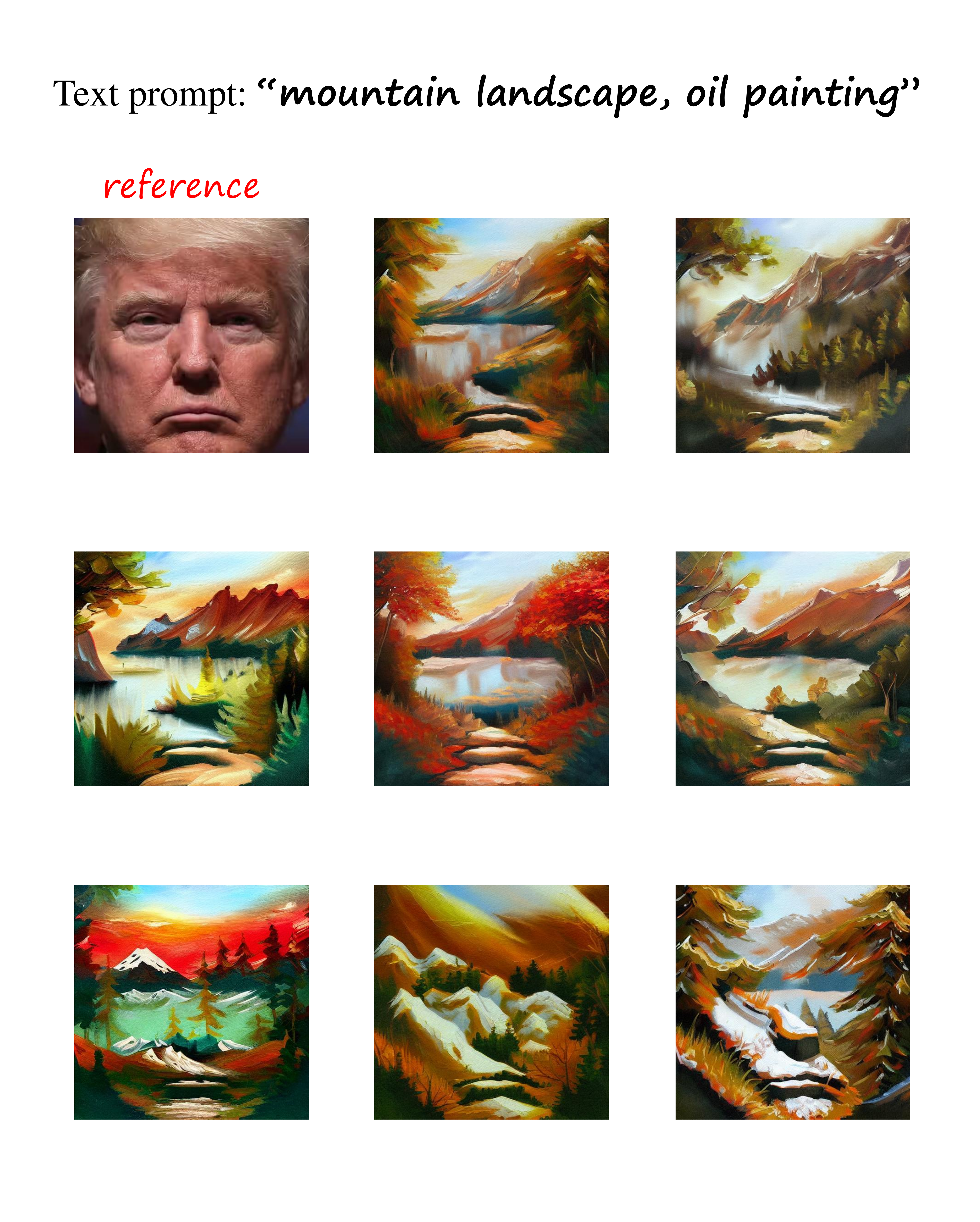}
    \caption{More examples of diversified sampling results of our method realized by varying the initial Gaussian noise $\tilde{z}_{T}$.}
    \label{fig:diverse_2}
\end{figure*}

\begin{figure*}[t]
    \centering
    \includegraphics[width=0.95\linewidth]{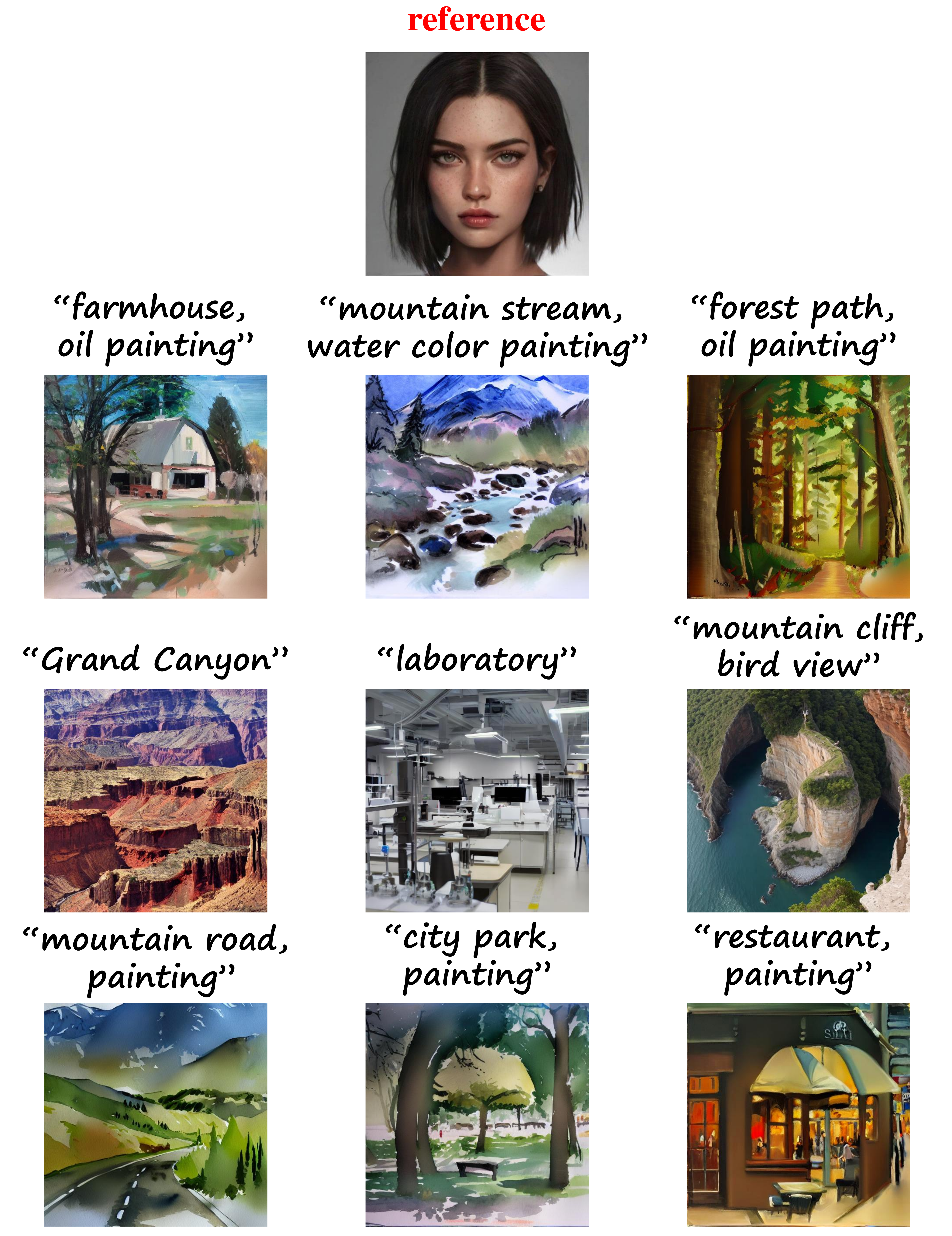}
    \caption{More qualitative results of our method.}
    \label{fig:case1}
\end{figure*}

\begin{figure*}[t]
    \centering
    \includegraphics[width=0.95\linewidth]{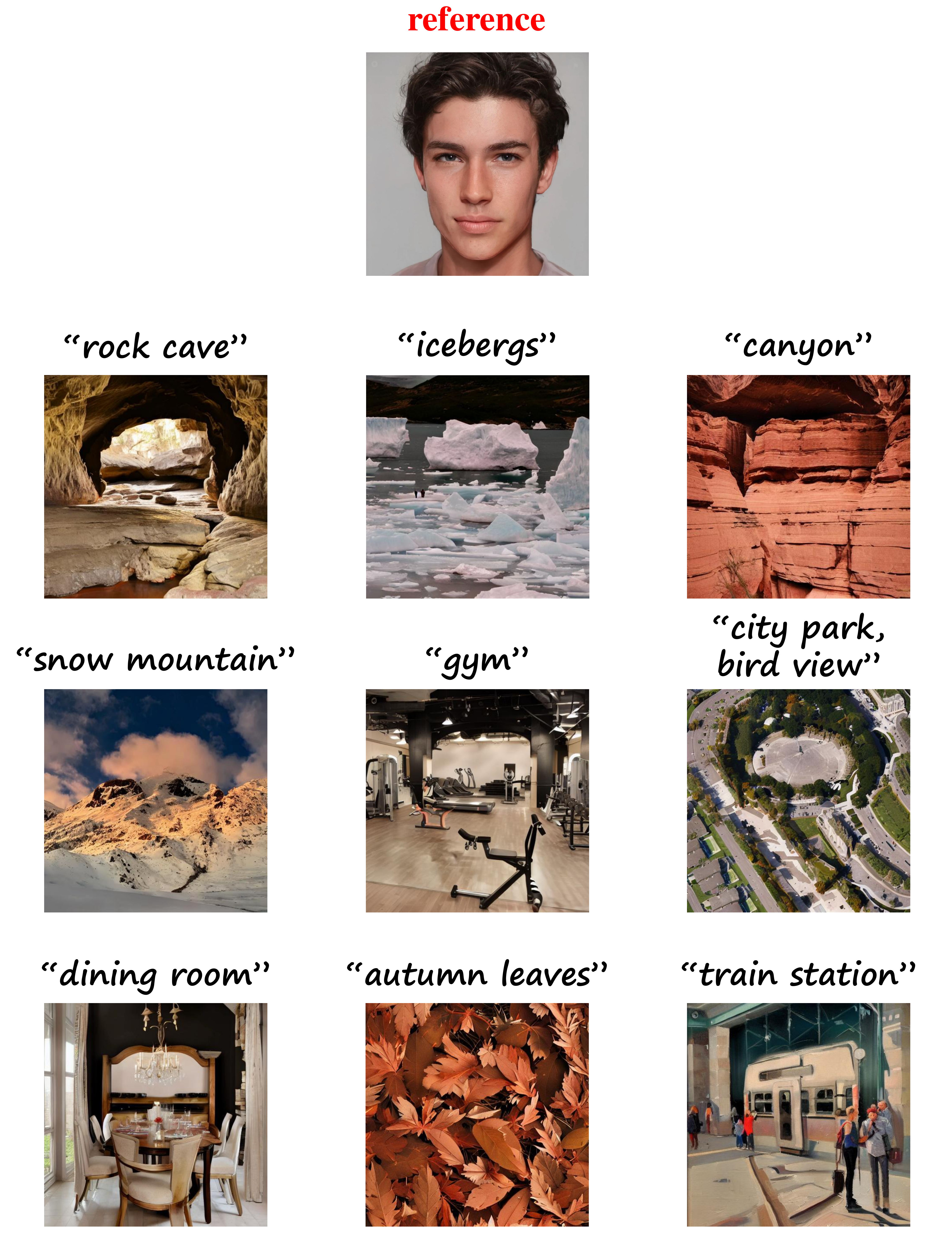}
    \caption{More qualitative results of our method.}
    \label{fig:case2}
\end{figure*}

\begin{figure*}[t]
    \centering
    \includegraphics[width=0.95\linewidth]{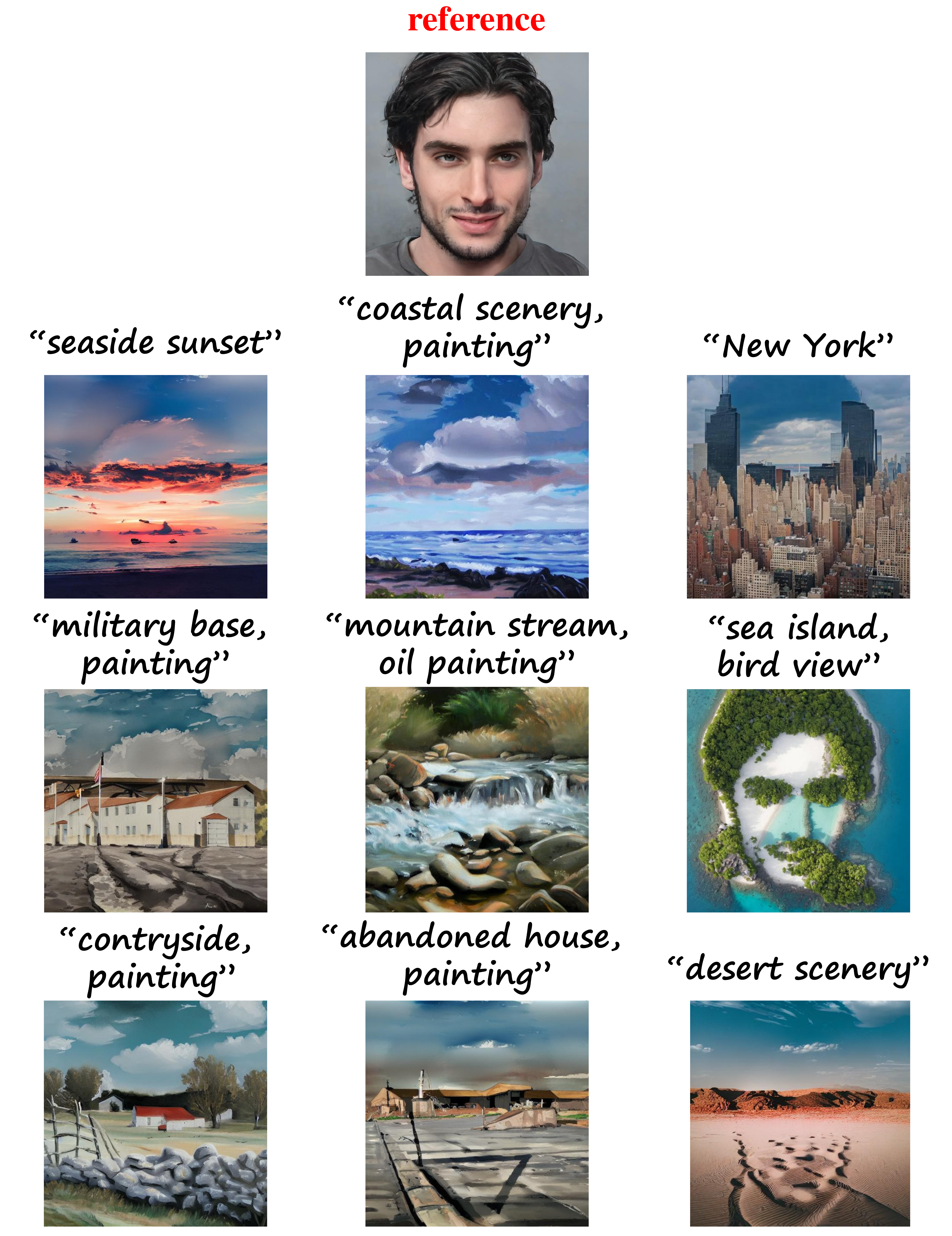}
    \caption{More qualitative results of our method.}
    \label{fig:case3}
\end{figure*}

\begin{figure*}[t]
    \centering
    \includegraphics[width=0.95\linewidth]{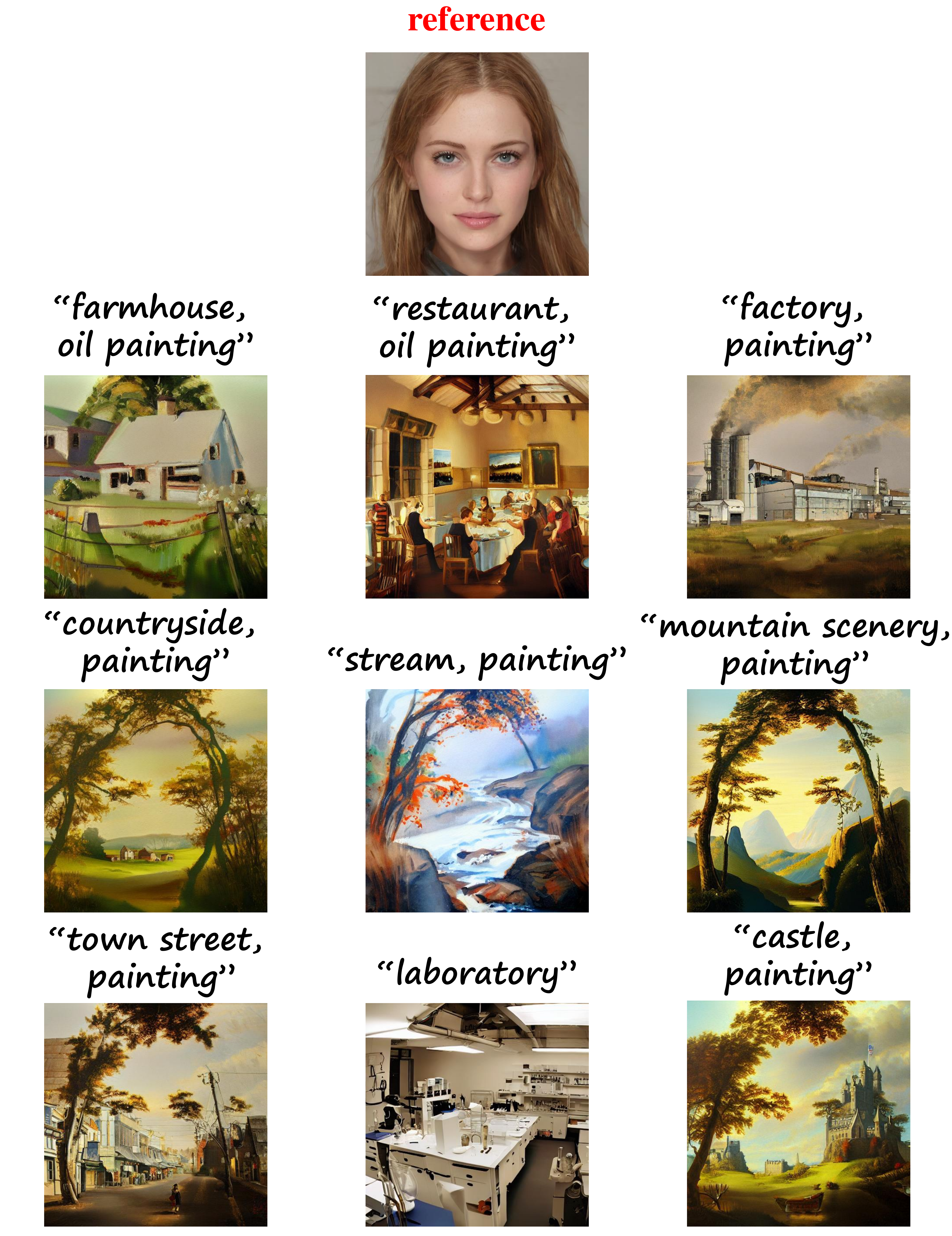}
    \caption{More qualitative results of our method.}
    \label{fig:case4}
\end{figure*}

\begin{figure*}[t]
    \centering
    \includegraphics[width=0.95\linewidth]{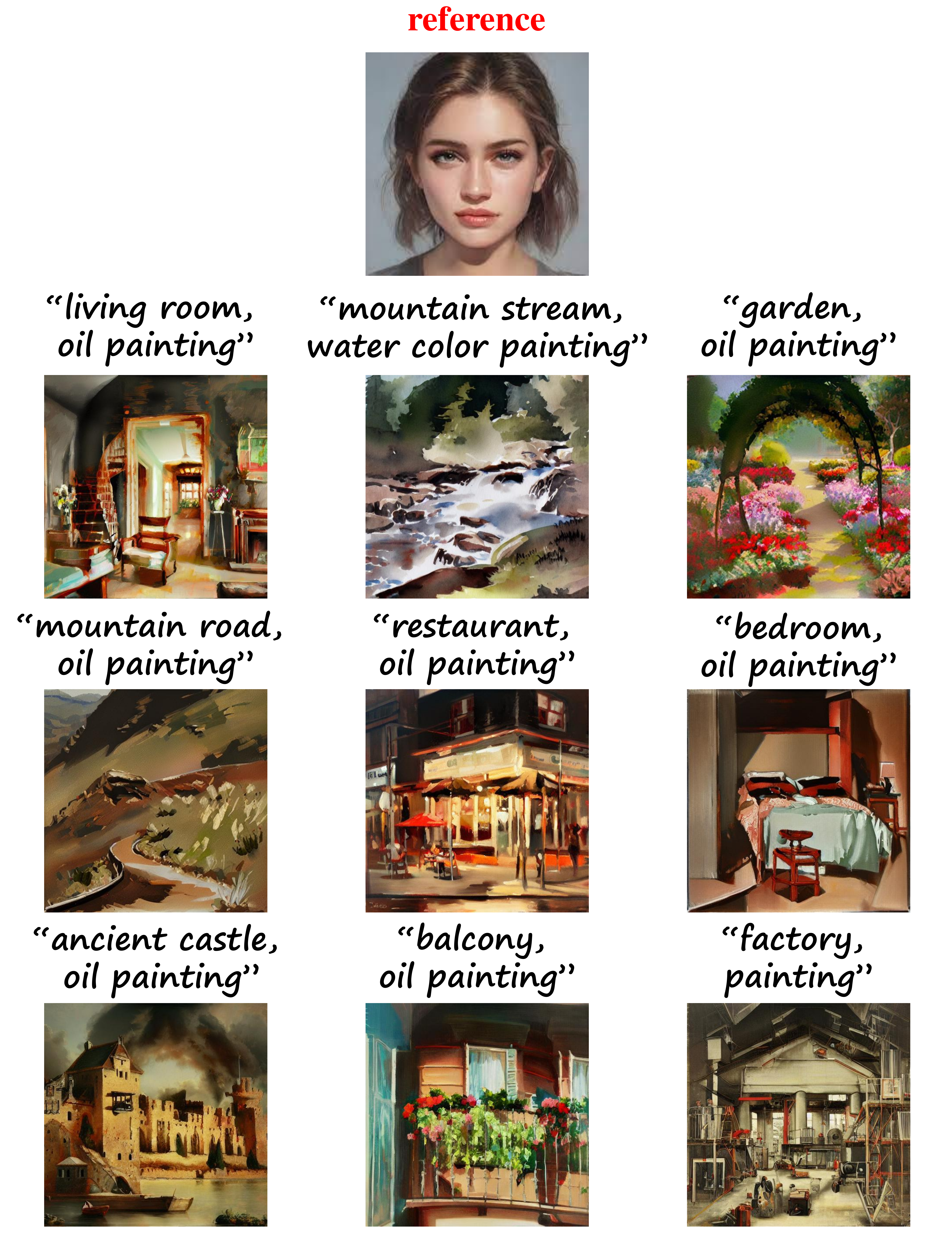}
    \caption{More qualitative results of our method.}
    \label{fig:case5}
\end{figure*}

\begin{figure*}[t]
    \centering
    \includegraphics[width=0.95\linewidth]{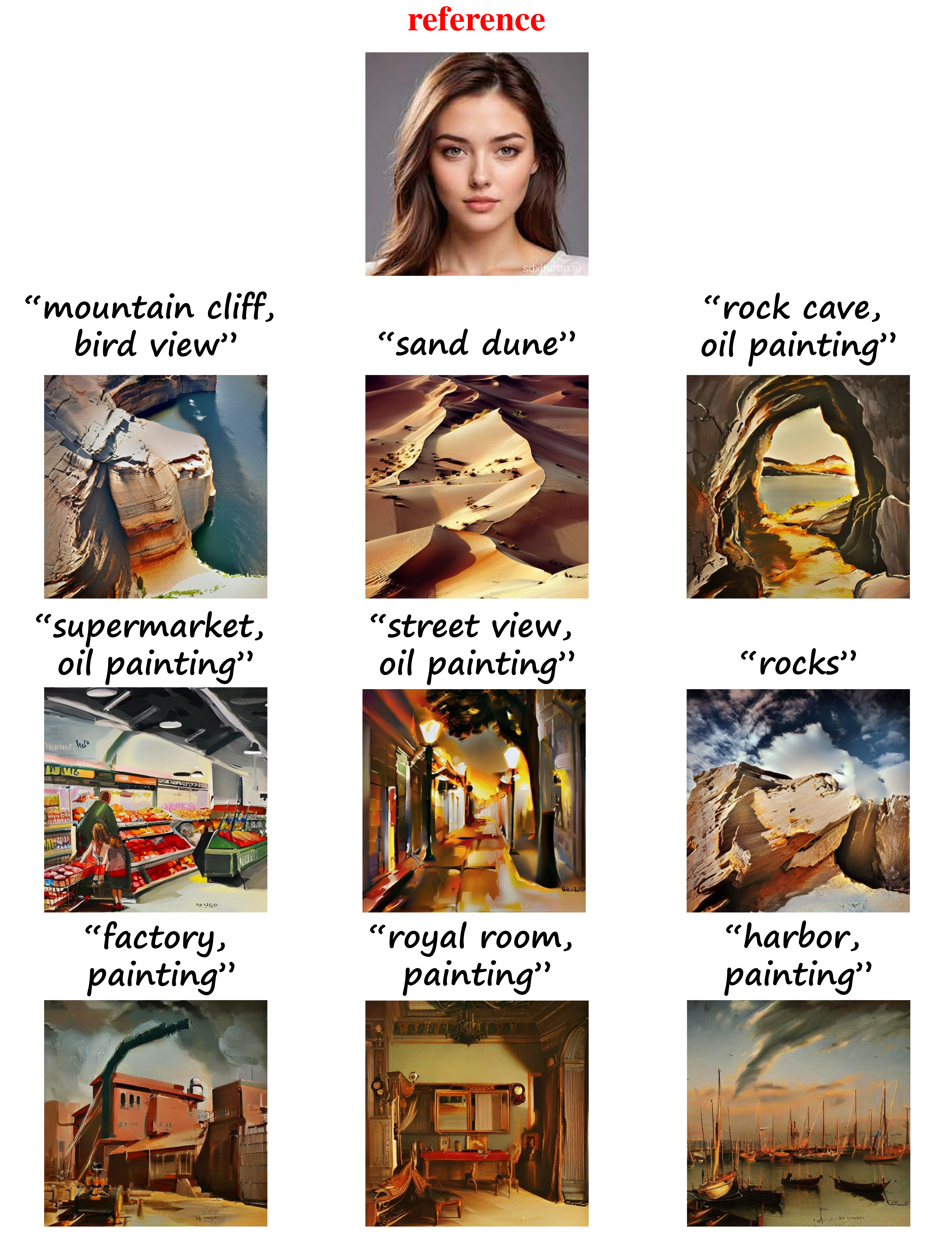}
    \caption{More qualitative results of our method.}
    \label{fig:case6}
\end{figure*}

\begin{figure*}[t]
    \centering
    \includegraphics[width=0.95\linewidth]{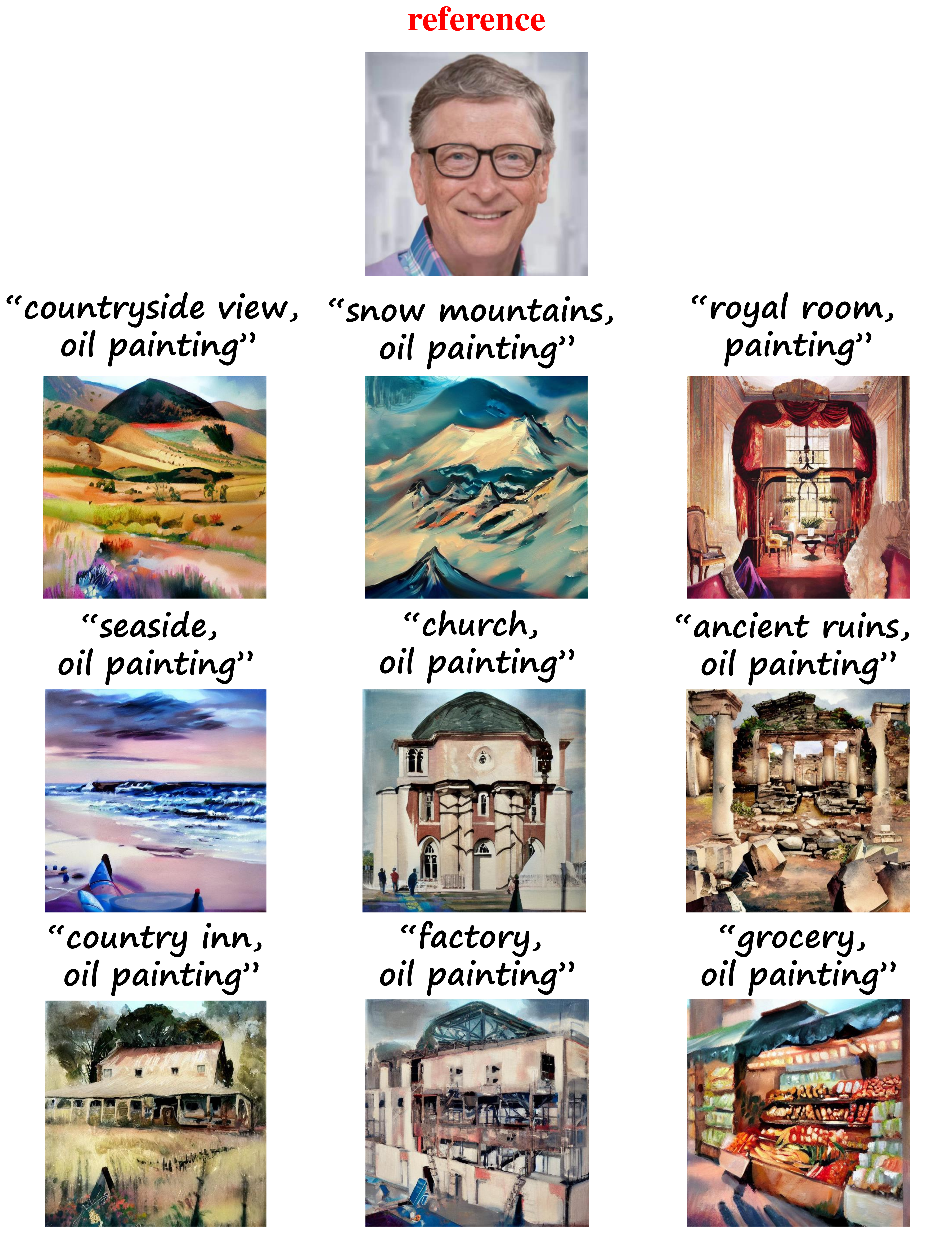}
    \caption{More qualitative results of our method.}
    \label{fig:case7}
\end{figure*}

\begin{figure*}[t]
    \centering
    \includegraphics[width=0.95\linewidth]{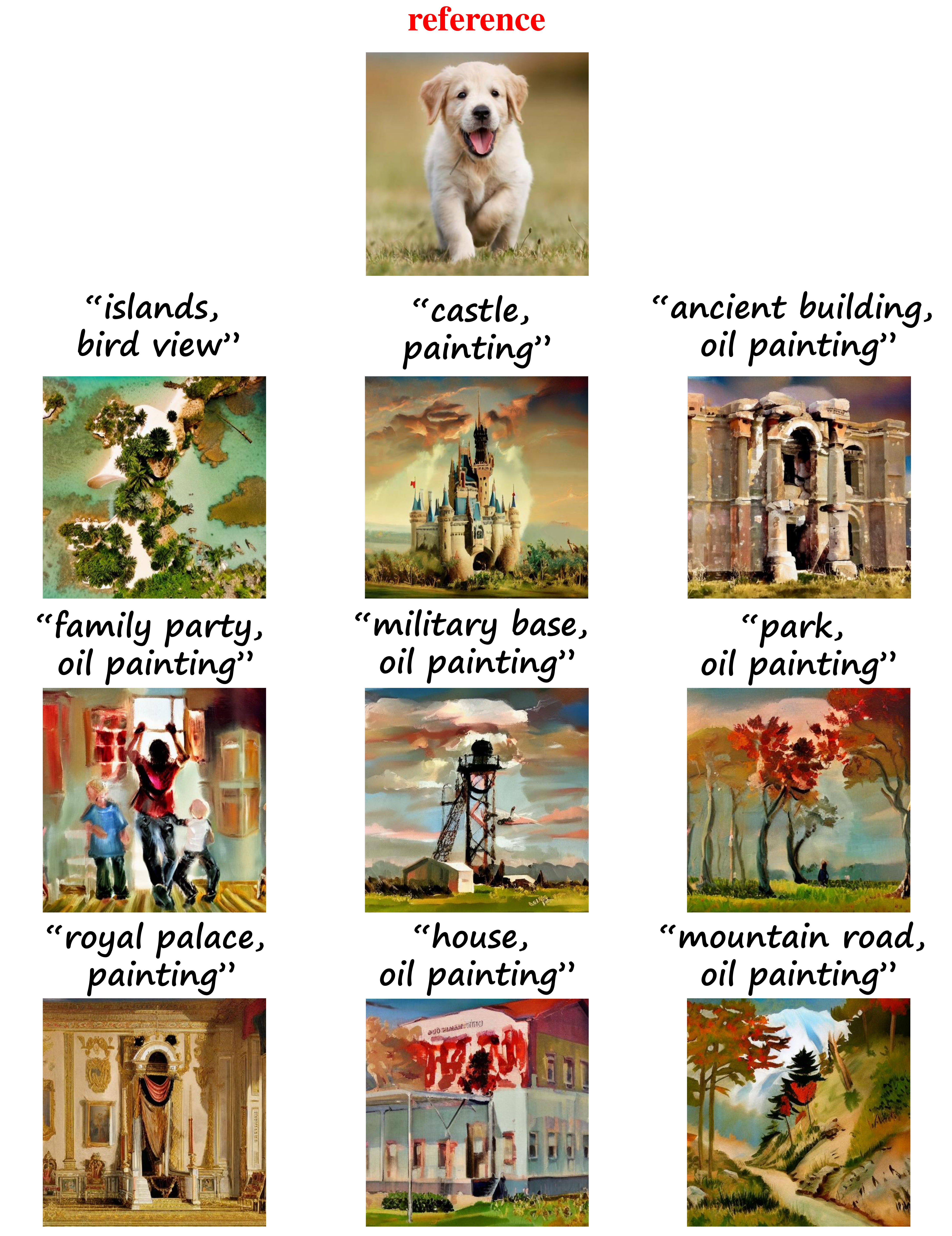}
    \caption{More qualitative results of our method.}
    \label{fig:case8}
\end{figure*}

\begin{figure*}[t]
    \centering
    \includegraphics[width=0.95\linewidth]{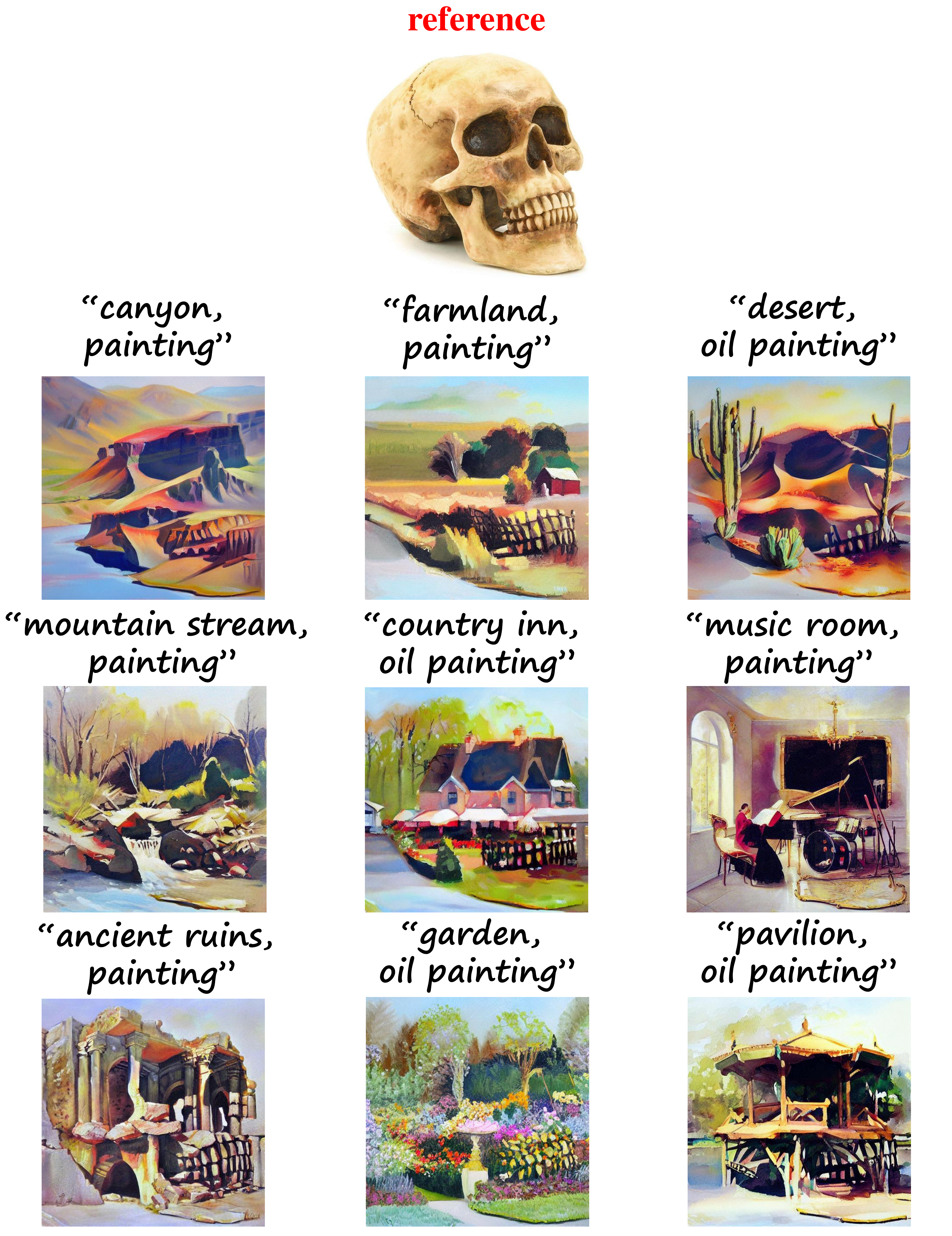}
    \caption{More qualitative results of our method.}
    \label{fig:case9}
\end{figure*}

\begin{figure*}[t]
    \centering
    \includegraphics[width=0.95\linewidth]{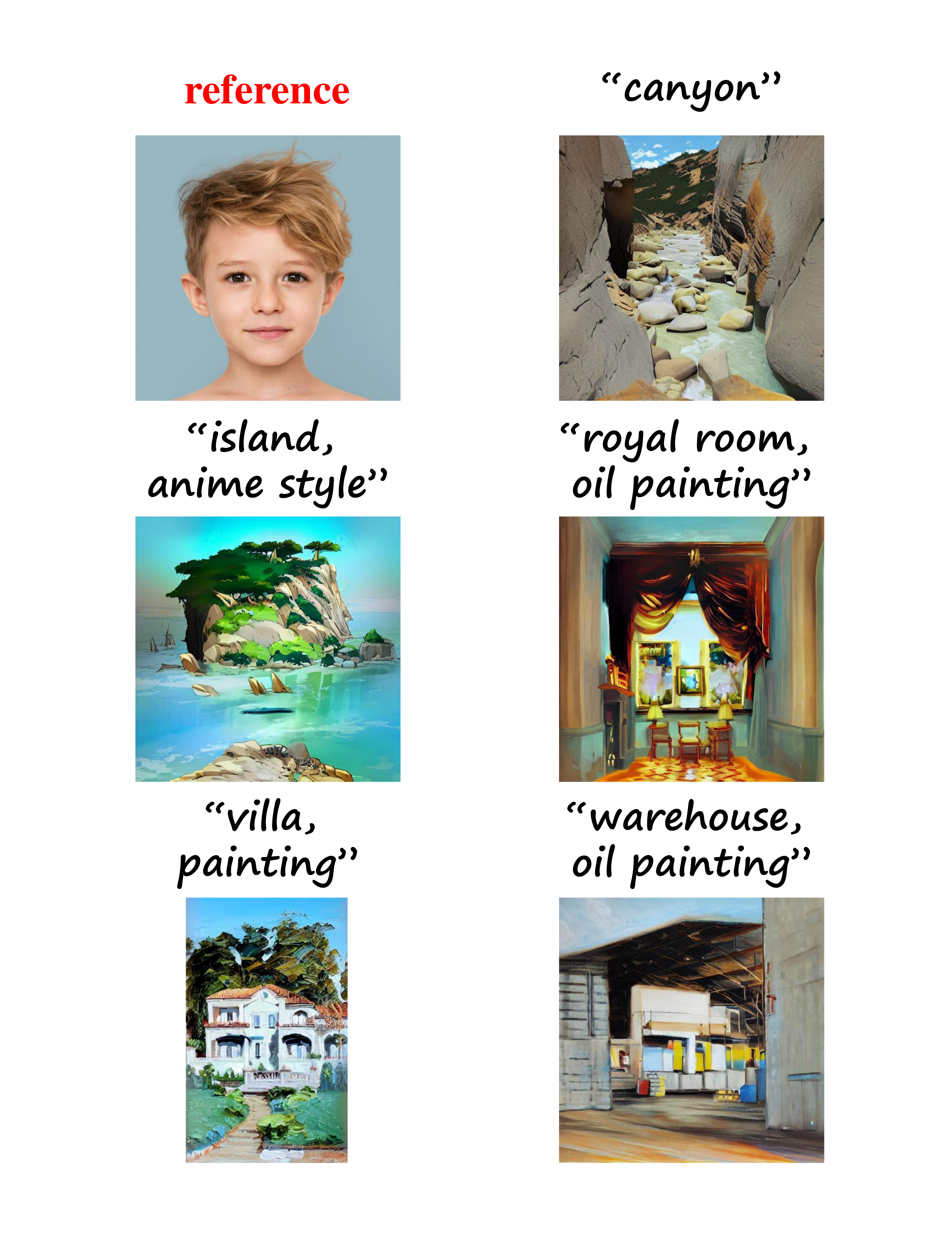}
    \caption{More qualitative results of our method.}
    \label{fig:case10}
\end{figure*}

\begin{figure*}[t]
    \centering
    \includegraphics[width=0.95\linewidth]{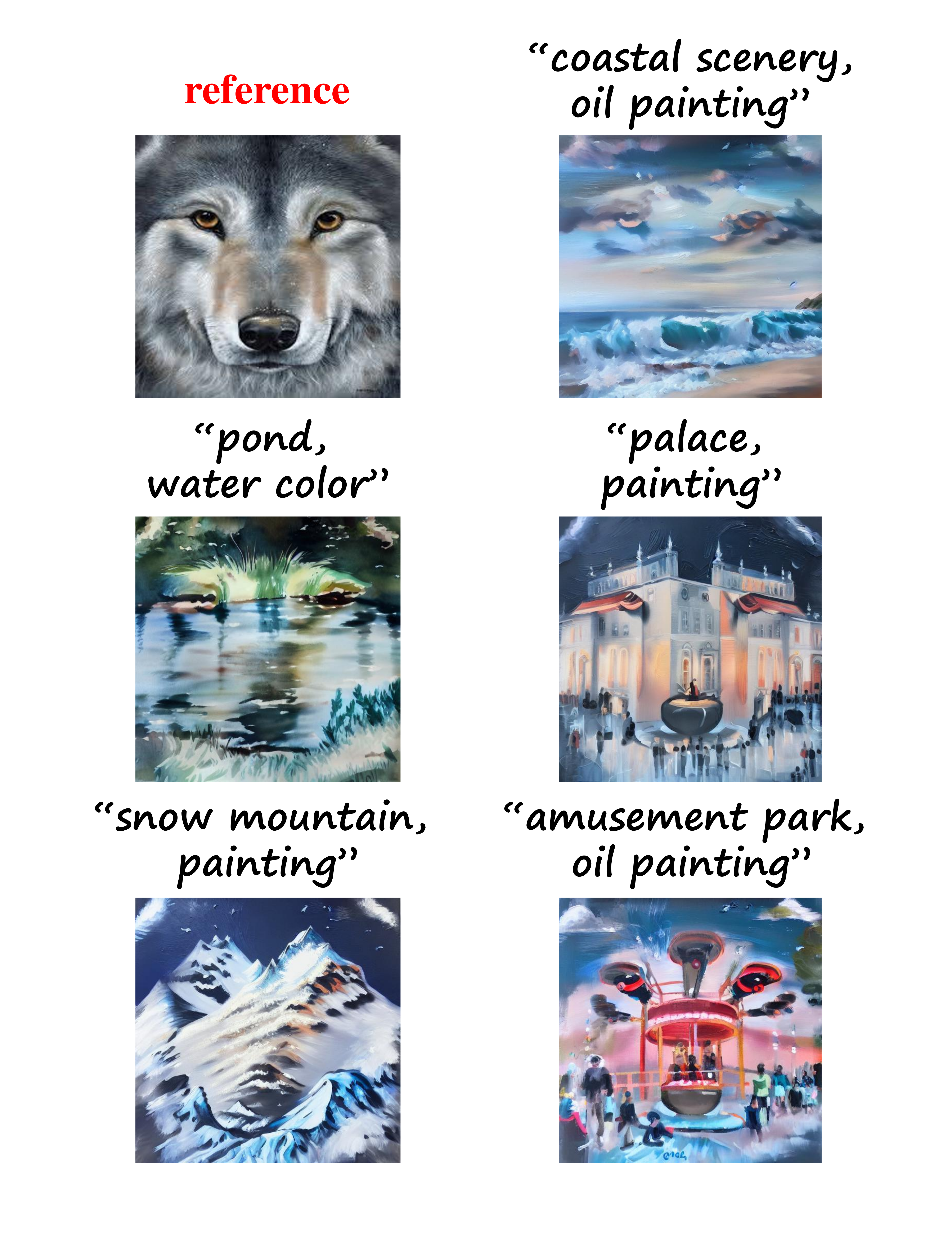}
    \caption{More qualitative results of our method.}
    \label{fig:case11}
\end{figure*}

Below we showcase more qualitative results of our PTDiffusion as a supplement to the main text. In Fig. \ref{fig:async_distance_1} and Fig. \ref{fig:async_distance_2}, we display more results of hidden content discernibility control realized by varying the async distance parameter $d$. In Fig. \ref{fig:diverse_1} and Fig. \ref{fig:diverse_2}, we display more examples demonstrating the sampling diversity property of our method, namely generating diversified illusion pictures with fixed reference image and text prompt. Finally, we present more optical illusion hidden pictures generated by our method in Fig. \ref{fig:case1} to Fig. \ref{fig:case11}.

{
    \small
    \bibliographystyle{ieeenat_fullname}
    \bibliography{main}

\begin{thebibliography}{51}
\providecommand{\natexlab}[1]{#1}
\providecommand{\url}[1]{\texttt{#1}}
\expandafter\ifx\csname urlstyle\endcsname\relax
  \providecommand{\doi}[1]{doi: #1}\else
  \providecommand{\doi}{doi: \begingroup \urlstyle{rm}\Url}\fi

\bibitem[Azadi et~al.(2018)Azadi, Fisher, Kim, Wang, Shechtman, and Darrell]{azadi2018multi}
Samaneh Azadi, Matthew Fisher, Vladimir~G Kim, Zhaowen Wang, Eli Shechtman, and Trevor Darrell.
\newblock Multi-content gan for few-shot font style transfer.
\newblock In \emph{Proceedings of the IEEE Conference on Computer Vision and Pattern Recognition}, pages 7564--7573, 2018.

\bibitem[Burgert et~al.(2024)Burgert, Li, Leite, Ranasinghe, and Ryoo]{burgert2024diffusion}
Ryan Burgert, Xiang Li, Abe Leite, Kanchana Ranasinghe, and Michael Ryoo.
\newblock Diffusion illusions: Hiding images in plain sight.
\newblock In \emph{Proceedings of the ACM SIGGRAPH}, pages 1--11, 2024.

\bibitem[Chu et~al.(2010)Chu, Hsu, Mitra, Cohen-Or, Wong, and Lee]{chu2010camouflage}
Hung-Kuo Chu, Wei-Hsin Hsu, Niloy~J Mitra, Daniel Cohen-Or, Tien-Tsin Wong, and Tong-Yee Lee.
\newblock Camouflage images.
\newblock In \emph{Proceedings of the ACM SIGGRAPH}, pages 1--8. 2010.

\bibitem[Dhariwal and Nichol(2021)]{dhariwal2021diffusion}
Prafulla Dhariwal and Alex Nichol.
\newblock Diffusion models beat gans on image synthesis.
\newblock In \emph{Proceedings of the Conference on Neural Information Processing Systems}, pages 8780--8794, 2021.

\bibitem[Dong et~al.(2023)Dong, Xue, Duan, and Han]{dong2023prompt}
Wenkai Dong, Song Xue, Xiaoyue Duan, and Shumin Han.
\newblock Prompt tuning inversion for text-driven image editing using diffusion models.
\newblock In \emph{Proceedings of the IEEE International Conference on Computer Vision}, pages 7430--7440, 2023.

\bibitem[Ehm(2011)]{ehm2011variational}
Werner Ehm.
\newblock A variational approach to geometric-optical illusions modeling.
\newblock \emph{Proceedings of Fechner Day}, 27\penalty0 (1):\penalty0 41--46, 2011.

\bibitem[Esser et~al.(2024)Esser, Kulal, Blattmann, Entezari, M{\"u}ller, Saini, Levi, Lorenz, Sauer, Boesel, et~al.]{esser2024scaling}
Patrick Esser, Sumith Kulal, Andreas Blattmann, Rahim Entezari, Jonas M{\"u}ller, Harry Saini, Yam Levi, Dominik Lorenz, Axel Sauer, Frederic Boesel, et~al.
\newblock Scaling rectified flow transformers for high-resolution image synthesis.
\newblock In \emph{Proceedings of the International Conference on Machine Learning}, 2024.

\bibitem[Freeman et~al.(1991)Freeman, Adelson, and Heeger]{freeman1991motion}
William~T Freeman, Edward~H Adelson, and David~J Heeger.
\newblock Motion without movement.
\newblock In \emph{Proceedings of the Annual Conference on Computer Graphics and Interactive Techniques}, pages 27--30, 1991.

\bibitem[Fu et~al.(2019)Fu, Gong, Wang, Batmanghelich, Zhang, and Tao]{fu2019geometry}
Huan Fu, Mingming Gong, Chaohui Wang, Kayhan Batmanghelich, Kun Zhang, and Dacheng Tao.
\newblock Geometry-consistent generative adversarial networks for one-sided unsupervised domain mapping.
\newblock In \emph{Proceedings of the IEEE Conference on Computer Vision and Pattern Recognition}, pages 2427--2436, 2019.

\bibitem[Gao and Liu(2024)]{gao2024fbsdiff}
Xiang Gao and Jiaying Liu.
\newblock Fbsdiff: Plug-and-play frequency band substitution of diffusion features for highly controllable text-driven image translation.
\newblock In \emph{Proceedings of the ACM International Conference on Multimedia}, pages 4101--4109, 2024.

\bibitem[Gao and Zhang(2025)]{gao2025sragan}
Xiang Gao and Yuqi Zhang.
\newblock Sragan: Saliency regularized and attended generative adversarial network for chinese ink-wash painting style transfer.
\newblock \emph{Pattern Recognition}, 162:\penalty0 111344, 2025.

\bibitem[Gao et~al.(2020)Gao, Tian, and Qi]{gao2020rpd}
Xiang Gao, Yingjie Tian, and Zhiquan Qi.
\newblock Rpd-gan: Learning to draw realistic paintings with generative adversarial network.
\newblock \emph{IEEE Transactions on Image Processing}, 29:\penalty0 8706--8720, 2020.

\bibitem[Gao et~al.(2022)Gao, Zhang, and Tian]{gao2022learning}
Xiang Gao, Yuqi Zhang, and Yingjie Tian.
\newblock Learning to incorporate texture saliency adaptive attention to image cartoonization.
\newblock In \emph{Proceedings of the International Conference on Machine Learning}, pages 7183--7207. PMLR, 2022.

\bibitem[Gao et~al.(2024)Gao, Xu, Zhao, and Liu]{gao2024frequency}
Xiang Gao, Zhengbo Xu, Junhan Zhao, and Jiaying Liu.
\newblock Frequency-controlled diffusion model for versatile text-guided image-to-image translation.
\newblock In \emph{Proceedings of the AAAI Conference on Artificial Intelligence}, pages 1824--1832, 2024.

\bibitem[Gatys et~al.(2016)Gatys, Ecker, and Bethge]{gatys2016image}
Leon~A Gatys, Alexander~S Ecker, and Matthias Bethge.
\newblock Image style transfer using convolutional neural networks.
\newblock In \emph{Proceedings of the IEEE Conference on Computer Vision and Pattern Recognition}, pages 2414--2423, 2016.

\bibitem[Geng et~al.(2024)Geng, Park, and Owens]{geng2024visual}
Daniel Geng, Inbum Park, and Andrew Owens.
\newblock Visual anagrams: Generating multi-view optical illusions with diffusion models.
\newblock In \emph{Proceedings of the IEEE Conference on Computer Vision and Pattern Recognition}, pages 24154--24163, 2024.

\bibitem[Goodfellow et~al.(2020)Goodfellow, Pouget-Abadie, Mirza, Xu, Warde-Farley, Ozair, Courville, and Bengio]{goodfellow2020generative}
Ian Goodfellow, Jean Pouget-Abadie, Mehdi Mirza, Bing Xu, David Warde-Farley, Sherjil Ozair, Aaron Courville, and Yoshua Bengio.
\newblock Generative adversarial networks.
\newblock \emph{Communications of the ACM}, 63\penalty0 (11):\penalty0 139--144, 2020.

\bibitem[Hertz et~al.(2023)Hertz, Mokady, Tenenbaum, Aberman, Pritch, and Cohen-or]{hertzprompt}
Amir Hertz, Ron Mokady, Jay Tenenbaum, Kfir Aberman, Yael Pritch, and Daniel Cohen-or.
\newblock Prompt-to-prompt image editing with cross-attention control.
\newblock In \emph{Proceedings of The International Conference on Learning Representations}, 2023.

\bibitem[Hirsch and Tal(2020)]{hirsch2020color}
Elad Hirsch and Ayellet Tal.
\newblock Color visual illusions: a statistics-based computational model.
\newblock In \emph{Proceedings of the Neural Information Processing Systems}, pages 9447--9458, 2020.

\bibitem[Ho and Salimans(2022)]{ho2022classifier}
Jonathan Ho and Tim Salimans.
\newblock Classifier-free diffusion guidance.
\newblock \emph{arXiv preprint arXiv:2207.12598}, 2022.

\bibitem[Ho et~al.(2020)Ho, Jain, and Abbeel]{ho2020denoising}
Jonathan Ho, Ajay Jain, and Pieter Abbeel.
\newblock Denoising diffusion probabilistic models.
\newblock In \emph{Proceedings of the Conference on Neural Information Processing Systems}, pages 6840--6851, 2020.

\bibitem[Isola et~al.(2017)Isola, Zhu, Zhou, and Efros]{isola2017image}
Phillip Isola, Jun-Yan Zhu, Tinghui Zhou, and Alexei~A Efros.
\newblock Image-to-image translation with conditional adversarial networks.
\newblock In \emph{Proceedings of the IEEE Conference on Computer Vision and Pattern Recognition}, pages 1125--1134, 2017.

\bibitem[Jiang et~al.(2019)Jiang, Lian, Tang, and Xiao]{jiang2019scfont}
Yue Jiang, Zhouhui Lian, Yingmin Tang, and Jianguo Xiao.
\newblock Scfont: Structure-guided chinese font generation via deep stacked networks.
\newblock In \emph{Proceedings of the AAAI conference on artificial intelligence}, pages 4015--4022, 2019.

\bibitem[Jiang et~al.(2023)Jiang, Jiang, Yang, and Loy]{jiang2023scenimefy}
Yuxin Jiang, Liming Jiang, Shuai Yang, and Chen~Change Loy.
\newblock Scenimefy: Learning to craft anime scene via semi-supervised image-to-image translation.
\newblock In \emph{Proceedings of the IEEE International Conference on Computer Vision}, pages 7357--7367, 2023.

\bibitem[Karras et~al.(2019)Karras, Laine, and Aila]{karras2019style}
Tero Karras, Samuli Laine, and Timo Aila.
\newblock A style-based generator architecture for generative adversarial networks.
\newblock In \emph{Proceedings of the IEEE Conference on Computer Vision and Pattern Recognition}, pages 4401--4410, 2019.

\bibitem[Kawar et~al.(2023)Kawar, Zada, Lang, Tov, Chang, Dekel, Mosseri, and Irani]{kawar2023imagic}
Bahjat Kawar, Shiran Zada, Oran Lang, Omer Tov, Huiwen Chang, Tali Dekel, Inbar Mosseri, and Michal Irani.
\newblock Imagic: Text-based real image editing with diffusion models.
\newblock In \emph{Proceedings of the IEEE Conference on Computer Vision and Pattern Recognition}, pages 6007--6017, 2023.

\bibitem[Kotovenko et~al.(2019)Kotovenko, Sanakoyeu, Ma, Lang, and Ommer]{kotovenko2019content}
Dmytro Kotovenko, Artsiom Sanakoyeu, Pingchuan Ma, Sabine Lang, and Bjorn Ommer.
\newblock A content transformation block for image style transfer.
\newblock In \emph{Proceedings of the IEEE Conference on Computer Vision and Pattern Recognition}, pages 10032--10041, 2019.

\bibitem[Lamdouar et~al.(2023)Lamdouar, Xie, and Zisserman]{lamdouar2023making}
Hala Lamdouar, Weidi Xie, and Andrew Zisserman.
\newblock The making and breaking of camouflage.
\newblock In \emph{Proceedings of the IEEE International Conference on Computer Vision}, pages 832--842, 2023.

\bibitem[Liao et~al.(2024)Liao, Wang, Wang, Peng, Weng, Chou, and Chen]{liao2024diffqrcoder}
Jia-Wei Liao, Winston Wang, Tzu-Sian Wang, Li-Xuan Peng, Ju-Hsuan Weng, Cheng-Fu Chou, and Jun-Cheng Chen.
\newblock {DiffQRCoder}: Diffusion-based aesthetic qr code generation with scanning robustness guided iterative refinement.
\newblock \emph{arXiv preprint arXiv:2409.06355}, 2024.

\bibitem[Meng et~al.(2021)Meng, He, Song, Song, Wu, Zhu, and Ermon]{meng2021sdedit}
Chenlin Meng, Yutong He, Yang Song, Jiaming Song, Jiajun Wu, Jun-Yan Zhu, and Stefano Ermon.
\newblock {SDEdit}: Guided image synthesis and editing with stochastic differential equations.
\newblock \emph{arXiv preprint arXiv:2108.01073}, 2021.

\bibitem[Mokady et~al.(2023)Mokady, Hertz, Aberman, Pritch, and Cohen-Or]{mokady2023null}
Ron Mokady, Amir Hertz, Kfir Aberman, Yael Pritch, and Daniel Cohen-Or.
\newblock Null-text inversion for editing real images using guided diffusion models.
\newblock In \emph{Proceedings of the IEEE Conference on Computer Vision and Pattern Recognition}, pages 6038--6047, 2023.

\bibitem[Mou et~al.(2024)Mou, Wang, Xie, Wu, Zhang, Qi, and Shan]{mou2024t2i}
Chong Mou, Xintao Wang, Liangbin Xie, Yanze Wu, Jian Zhang, Zhongang Qi, and Ying Shan.
\newblock {T2I-Adapter}: Learning adapters to dig out more controllable ability for text-to-image diffusion models.
\newblock In \emph{Proceedings of the AAAI Conference on Artificial Intelligence}, pages 4296--4304, 2024.

\bibitem[Nichol et~al.(2022)Nichol, Dhariwal, Ramesh, Shyam, Mishkin, Mcgrew, Sutskever, and Chen]{nichol2022glide}
Alexander~Quinn Nichol, Prafulla Dhariwal, Aditya Ramesh, Pranav Shyam, Pamela Mishkin, Bob Mcgrew, Ilya Sutskever, and Mark Chen.
\newblock {GLIDE}: Towards photorealistic image generation and editing with text-guided diffusion models.
\newblock In \emph{Proceedings of the International Conference on Machine Learning}, pages 16784--16804. PMLR, 2022.

\bibitem[Oliva et~al.(2006)Oliva, Torralba, and Schyns]{oliva2006hybrid}
Aude Oliva, Antonio Torralba, and Philippe~G Schyns.
\newblock Hybrid images.
\newblock \emph{ACM Transactions on Graphics}, 25\penalty0 (3):\penalty0 527--532, 2006.

\bibitem[Park et~al.(2020)Park, Efros, Zhang, and Zhu]{park2020contrastive}
Taesung Park, Alexei~A Efros, Richard Zhang, and Jun-Yan Zhu.
\newblock Contrastive learning for unpaired image-to-image translation.
\newblock In \emph{Proceedings of the European Conference on Computer Vision}, pages 319--345. Springer, 2020.

\bibitem[Parmar et~al.(2023)Parmar, Kumar~Singh, Zhang, Li, Lu, and Zhu]{parmar2023zero}
Gaurav Parmar, Krishna Kumar~Singh, Richard Zhang, Yijun Li, Jingwan Lu, and Jun-Yan Zhu.
\newblock Zero-shot image-to-image translation.
\newblock In \emph{Proceedings of the ACM SIGGRAPH}, pages 1--11, 2023.

\bibitem[Podell et~al.(2023)Podell, English, Lacey, Blattmann, Dockhorn, M{\"u}ller, Penna, and Rombach]{podell2023sdxl}
Dustin Podell, Zion English, Kyle Lacey, Andreas Blattmann, Tim Dockhorn, Jonas M{\"u}ller, Joe Penna, and Robin Rombach.
\newblock {SDXL}: Improving latent diffusion models for high-resolution image synthesis.
\newblock \emph{arXiv preprint arXiv:2307.01952}, 2023.

\bibitem[Poole et~al.(2022)Poole, Jain, Barron, and Mildenhall]{poole2022dreamfusion}
Ben Poole, Ajay Jain, Jonathan~T Barron, and Ben Mildenhall.
\newblock {DreamFusion}: Text-to-{3D} using {2D} diffusion.
\newblock \emph{arXiv preprint arXiv:2209.14988}, 2022.

\bibitem[Ramesh et~al.(2022)Ramesh, Dhariwal, Nichol, Chu, and Chen]{ramesh2022hierarchical}
Aditya Ramesh, Prafulla Dhariwal, Alex Nichol, Casey Chu, and Mark Chen.
\newblock Hierarchical text-conditional image generation with {CLIP} latents.
\newblock \emph{arXiv preprint arXiv:2204.06125}, 2022.

\bibitem[Rombach et~al.(2022)Rombach, Blattmann, Lorenz, Esser, and Ommer]{rombach2022high}
Robin Rombach, Andreas Blattmann, Dominik Lorenz, Patrick Esser, and Bj{\"o}rn Ommer.
\newblock High-resolution image synthesis with latent diffusion models.
\newblock In \emph{Proceedings of the IEEE Conference on Computer Vision and Pattern Recognition}, pages 10684--10695, 2022.

\bibitem[Saharia et~al.(2022{\natexlab{a}})Saharia, Chan, Chang, Lee, Ho, Salimans, Fleet, and Norouzi]{saharia2022palette}
Chitwan Saharia, William Chan, Huiwen Chang, Chris Lee, Jonathan Ho, Tim Salimans, David Fleet, and Mohammad Norouzi.
\newblock Palette: Image-to-image diffusion models.
\newblock In \emph{Proceedings of the ACM SIGGRAPH}, pages 1--10, 2022{\natexlab{a}}.

\bibitem[Saharia et~al.(2022{\natexlab{b}})Saharia, Chan, Saxena, Lit, Whang, Denton, Ghasemipour, Ayan, Mahdavi, Gontijo-Lopes, et~al.]{saharia2022photorealistic}
Chitwan Saharia, William Chan, Saurabh Saxena, Lala Lit, Jay Whang, Emily Denton, Seyed Kamyar~Seyed Ghasemipour, Burcu~Karagol Ayan, S~Sara Mahdavi, Raphael Gontijo-Lopes, et~al.
\newblock Photorealistic text-to-image diffusion models with deep language understanding.
\newblock In \emph{Proceedings of the Conference on Neural Information Processing Systems}, pages 36479--36494, 2022{\natexlab{b}}.

\bibitem[Song et~al.(2020)Song, Meng, and Ermon]{song2020denoising}
Jiaming Song, Chenlin Meng, and Stefano Ermon.
\newblock Denoising diffusion implicit models.
\newblock \emph{arXiv preprint arXiv:2010.02502}, 2020.

\bibitem[Tumanyan et~al.(2022)Tumanyan, Bar-Tal, Bagon, and Dekel]{tumanyan2022splicing}
Narek Tumanyan, Omer Bar-Tal, Shai Bagon, and Tali Dekel.
\newblock Splicing vit features for semantic appearance transfer.
\newblock In \emph{Proceedings of the IEEE Conference on Computer Vision and Pattern Recognition}, pages 10748--10757, 2022.

\bibitem[Tumanyan et~al.(2023)Tumanyan, Geyer, Bagon, and Dekel]{tumanyan2023plug}
Narek Tumanyan, Michal Geyer, Shai Bagon, and Tali Dekel.
\newblock Plug-and-play diffusion features for text-driven image-to-image translation.
\newblock In \emph{Proceedings of the IEEE Conference on Computer Vision and Pattern Recognition}, pages 1921--1930, 2023.

\bibitem[Wang et~al.(2023)Wang, Du, Li, Yeh, and Shakhnarovich]{wang2023score}
Haochen Wang, Xiaodan Du, Jiahao Li, Raymond~A Yeh, and Greg Shakhnarovich.
\newblock Score jacobian chaining: Lifting pretrained {2D} diffusion models for {3D} generation.
\newblock In \emph{Proceedings of the IEEE Conference on Computer Vision and Pattern Recognition}, pages 12619--12629, 2023.

\bibitem[Wu et~al.(2024)Wu, Liu, Jia, Cui, and Zhai]{wu2024text2qr}
Guangyang Wu, Xiaohong Liu, Jun Jia, Xuehao Cui, and Guangtao Zhai.
\newblock {Text2QR}: Harmonizing aesthetic customization and scanning robustness for text-guided {QR} code generation.
\newblock In \emph{Proceedings of the IEEE Conference on Computer Vision and Pattern Recognition}, pages 8456--8465, 2024.

\bibitem[Zhang et~al.(2023)Zhang, Rao, and Agrawala]{zhang2023adding}
Lvmin Zhang, Anyi Rao, and Maneesh Agrawala.
\newblock Adding conditional control to text-to-image diffusion models.
\newblock In \emph{Proceedings of the IEEE International Conference on Computer Vision}, pages 3836--3847, 2023.

\bibitem[Zhang et~al.(2020)Zhang, Yin, Nie, and Zheng]{zhang2020deep}
Qing Zhang, Gelin Yin, Yongwei Nie, and Wei-Shi Zheng.
\newblock Deep camouflage images.
\newblock In \emph{Proceedings of the AAAI Conference on Artificial Intelligence}, pages 12845--12852, 2020.

\bibitem[Zhao et~al.(2023)Zhao, Chen, Chen, Bao, Hao, Yuan, and Wong]{zhao2023uni}
Shihao Zhao, Dongdong Chen, Yen-Chun Chen, Jianmin Bao, Shaozhe Hao, Lu Yuan, and Kwan-Yee~K Wong.
\newblock {Uni-ControlNet}: all-in-one control to text-to-image diffusion models.
\newblock In \emph{Proceedings of the Conference on Neural Information Processing Systems}, pages 11127--11150, 2023.

\bibitem[Zhu et~al.(2017)Zhu, Park, Isola, and Efros]{zhu2017unpaired}
Jun-Yan Zhu, Taesung Park, Phillip Isola, and Alexei~A Efros.
\newblock Unpaired image-to-image translation using cycle-consistent adversarial networks.
\newblock In \emph{Proceedings of the IEEE International Conference on Computer Vision}, pages 2223--2232, 2017.

\end{thebibliography}
}


\end{document}